%% file: main.tex
\documentclass[conference]{IEEEtran}
\IEEEoverridecommandlockouts

\input{packages}
\input{commands}

\begin{document}

\input{title}
\maketitle

\input{abstract}
\input{introduction}
\input{related_work}
\input{method}

\input{experiments}
\input{conclusion}

\bibliographystyle{IEEEtran}
\bibliography{references}
\clearpage
\input{appendix}

\end{document}

%% file: packages.tex
\usepackage{cite}
\usepackage{amsmath,amssymb,amsfonts}
\usepackage{algorithmic}
\usepackage{graphicx}
\usepackage{textcomp}
\usepackage{listings}
\usepackage{subcaption}
\usepackage{xcolor}
\usepackage{gensymb}
\usepackage{booktabs}
\usepackage{url}
\usepackage{multirow}
\graphicspath{{figs/}}
\usepackage{iftex}
\usepackage{colortbl}
\ifPDFTeX
\else
  \usepackage{fontspec}
\fi

\lstdefinestyle{promptstyle}{
    basicstyle=\scriptsize\ttfamily,
    breaklines=true,
    breakatwhitespace=true,
    breakindent=0pt,
    columns=flexible,
    keepspaces=true,
    showstringspaces=false,
    frame=none,
    xleftmargin=0pt,
    xrightmargin=0pt,
    aboveskip=4pt,
    belowskip=4pt,
    postbreak={},
}

\definecolor{winbg}{RGB}{220,240,220}    
\definecolor{lossbg}{RGB}{250,220,220}   

%% file: commands.tex
\def\BibTeX{{\rm B\kern-.05em{\sc i\kern-.025em b}\kern-.08em
    T\kern-.1667em\lower.7ex\hbox{E}\kern-.125emX}}

\definecolor{dypink}{HTML}{ec008c}
\definecolor{dypurple}{HTML}{8654d1}

\newif\ifnotes
\notestrue 



%% file: title.tex
\newcommand{\sys}{ConceptTS}


\title{\sys: LLM-Guided Concept Bottlenecks for Interpretable Multivariate Time-Series Forecasting
}

\author{\IEEEauthorblockN{1\textsuperscript{st} Yichen Jiang\IEEEauthorrefmark{1}\thanks{\IEEEauthorrefmark{1}\,Yichen Jiang joined this VIA Lab project at UC Davis while affiliated with Stanford University and later completed his work as a remote research intern under Dongyu Liu's supervision.}}
\IEEEauthorblockA{\textit{Department of Electrical Engineering} \\
\textit{Stanford University}\\
Stanford, CA, USA\\
ycjiang@stanford.edu}
\and
\IEEEauthorblockN{2\textsuperscript{nd} Yueqiao Chen}
\IEEEauthorblockA{\textit{Department of Computer Science} \\
\textit{University of California, Davis}\\
Davis, CA, USA\\
yeqchen@ucdavis.edu}
\and
\IEEEauthorblockN{3\textsuperscript{rd} Dongyu Liu}
\IEEEauthorblockA{\textit{Department of Computer Science} \\
\textit{University of California, Davis}\\
Davis, CA, USA\\
dyuliu@ucdavis.edu}
}

\maketitle

%% file: abstract.tex
\begin{abstract}
State-of-the-art multivariate time-series forecasters can model complex temporal and cross-variable dependencies, yet their opaque representations provide limited insight into why a particular forecast is produced. This lack of transparency restricts their use in settings where practitioners must understand and assess the factors underlying a prediction. We introduce \sys, an interpretable forecasting framework that organizes its predictions around named, human-readable concepts. \sys~uses a large language model to propose task-relevant concepts and generate executable labeling rules, translating the language model's domain knowledge into direct supervision without costly manual concept annotation. The proposed concepts are organized into three complementary bottlenecks that describe the historical context, local forecast intervals, and the full forecast horizon. A shared decoder combines representations derived from their predicted activations to construct the forecast, making the model's decision process explicit and supporting direct concept-level interventions. Experiments on the Beijing Multi-Site Air Quality dataset show that \sys~achieves accuracy competitive with strong black-box baselines while producing semantically meaningful concept activations.
\end{abstract}


%% file: introduction.tex
\section{Introduction}

Multivariate time series forecasting predicts a target trajectory from historical observations and correlated covariates. Across healthcare, environmental monitoring, energy, and industry, such continuously collected data demand both scalable modeling and interpretable temporal analysis~\cite{liu2022mtv,huang2026sigtime}. Modern recurrent, convolutional, graph, linear/MLP, and Transformer architectures can capture nonlinear temporal dynamics and cross-variable dependencies with strong accuracy~\cite{wang2026deep}. Yet their predictive progress has outpaced interpretability: a forecast does not reveal which temporal conditions or variable interactions shaped it, or whether the model used plausible evidence. This opacity is consequential in high-stakes, data-intensive settings, where practitioners must understand potential failure modes before trusting a prediction.

Interpreting multivariate time series is particularly challenging because predictive evidence is distributed across variables and temporal scales, and its relevance can change throughout the forecast horizon~\cite{theissler2022explainable,zhao2023interpretation,park2026transparent}. A forecast may depend jointly on a long-term trend in one channel, a short-lived event in another, and their evolving interaction. Attribution or saliency methods can identify influential channels and timestamps, but these low-level importance scores do not necessarily explain the semantic condition represented by the selected evidence. Concept-based models offer a promising alternative by organizing predictions around human-understandable intermediate variables. Concept bottleneck models (CBMs) make these variables inspectable and directly intervenable~\cite{koh2020conceptbottleneckmodels}, while recent time-series methods provide complementary directions: ProtoTS introduces hierarchical prototype reasoning for forecasting~\cite{peng2025protots}, and TimeX++ learns label-preserving explanatory instances through an information bottleneck~\cite{liu2024timexplusplus}. Nevertheless, conventional CBMs depend on expert-defined concepts and per-example annotations; ProtoTS learns latent prototypes whose meanings must be inferred after training; and TimeX++ explains an existing predictor rather than requiring forecasts to pass through named semantic evidence. Thus, it remains difficult to obtain an accurate forecaster whose internal evidence is semantically explicit, supervised, and tied to the prediction pathway without incurring substantial annotation cost.

We address this gap with \textbf{\sys}, an LLM-guided concept-bottleneck framework for interpretable multivariate time-series forecasting. Given domain context, dataset statistics, and representative temporal regimes, an offline large language model proposes concept names, natural-language definitions, and executable predicates that automatically label training segments. \sys~organizes these concepts into complementary bottlenecks describing the historical context, local intervals within the forecast horizon, and the horizon as a whole. A structure-preserving encoder predicts their activations, and a shared decoder constructs the forecast from the resulting concept representations. Input-independent concept embeddings limit information leakage around the displayed activation pathway, while a small regularized residual channel accommodates continuous magnitude information not fully represented by discrete concepts and exposes an explicit accuracy--interpretability trade-off. The LLM is used only during concept construction and is not required at inference time. We evaluate \sys~on three sites from the Beijing Multi-Site Air Quality dataset under both future-aware and look-back-only settings. The model remains competitive with strong black-box forecasters, including an MAE within $5\%$ of the best tested look-back-only baseline. Beyond predictive accuracy, concept-count and residual-channel ablations, qualitative case studies, counterfactual concept sweeps, and activation interventions examine how the concepts are learned and used. In particular, replacing the activations with inverted ground-truth concept labels increases test MAE from $15.371$ to $99.661$, providing direct evidence that the decoder relies on the concept-activation channel in the tested~configuration.

Our contributions are threefold:
\begin{itemize}
    \setlength{\itemsep}{2pt}
    \setlength{\parskip}{0pt}
    \item \textbf{LLM-guided concept construction and supervision.}
    We introduce a pipeline that converts domain context and dataset summaries into named temporal concepts, natural-language definitions, and executable labeling rules, avoiding manual concept annotation for individual \mbox{training~segments}.

    \item \textbf{A multiscale, intervention-ready forecasting architecture.}
    We organize concepts across historical, local-forecast, and global-forecast scopes and design a prediction pathway that preserves temporal and variable structure while limiting information leakage around the concept bottlenecks.

    \item \textbf{Evaluation of both forecasting and interpretability.}
    Experiments across three monitoring sites and two input regimes establish competitive forecasting performance, while ablations, case studies, concept sweeps, and controlled interventions evaluate concept quality, model reliance, and the accuracy--interpretability trade-off.
\end{itemize}

%% file: related_work.tex
\section{Related Work}

\subsection{Explainability for Time-Series Models}

Explaining time-series predictions requires more than assigning importance to independent features. Predictive evidence may lie in an event, its duration, a trend over a window, or an interaction among channels; moreover, the relevant evidence may change across the forecast horizon. Surveys therefore distinguish post-hoc attribution, perturbation, and example-based explanations from intrinsically interpretable architectures, and assess explanations along dimensions such as local versus global scope, faithfulness, stability, and human comprehensibility~\cite{theissler2022explainable,zhao2023interpretation,park2026transparent}. These requirements are particularly demanding for multivariate forecasting: correlated variables make individual attribution ambiguous, temporal aggregation can hide short-lived events, and masking may create out-of-distribution inputs~\cite{liu2024timexplusplus}.

Several forecasting methods expose which parts of an input appear influential. Series Saliency jointly learns a forecast and saliency over multivariate input regions, while the Temporal Fusion Transformer provides variable-selection weights and temporal attention patterns~\cite{Panetal2021,lim2020temporalfusiontransformersinterpretable}. TimeX++ addresses limitations of conventional masking by learning in-distribution, label-preserving explanatory instances through an information-bottleneck objective~\cite{liu2024timexplusplus}. These approaches provide useful localization and model diagnostics, but their explanations are primarily expressed as importance over variables and timestamps or as selected input instances. A practitioner must still infer the semantic condition represented by that evidence---for example, whether a joint change denotes a pollution episode, a weather transition, or a recurring daily regime. In addition, an explanation selected for an existing predictor is not necessarily the intermediate representation through which its forecast is computed. Our work instead targets an intrinsic forecasting pathway whose evidence is expressed in a stable vocabulary of named temporal concepts and can be directly intervened~on.

\subsection{Concept-Based and Language-Guided Models}

Concept bottleneck models were originally developed for image classification, where a model predicts human-defined visual attributes before using them to infer the target label~\cite{koh2020conceptbottleneckmodels}. Because the task prediction is expressed through these intermediate concepts, users can inspect and intervene on the model's reasoning. Concept embedding models increase this bottleneck's capacity by representing the active and inactive states of each concept with learned embeddings~\cite{espinosa2022cem}. Adapting this paradigm to multivariate time series is nontrivial: unlike static visual attributes, temporal concepts may describe trends, durations, cross-variable relationships, and patterns at different time scales. Moreover, defining these concepts and labeling every sequence normally requires substantial domain expertise. Our work adapts the concept-bottleneck paradigm to forecasting while reducing this supervision burden.

Language-guided methods offer a way to reduce this supervision burden. LaBo uses a language model to propose candidate concepts and grounds them through a vision-language model for interpretable image classification, while concept-bottleneck LLMs organize language-model decisions around human-readable intermediate concepts~\cite{yang2023languagebottlelanguagemodel,sun2025conceptbottlenecklargelanguage}. These studies demonstrate that language models can assist concept discovery, but they primarily address image or text tasks. Continuous multivariate sequences additionally require operational definitions that specify when a temporal concept is active, accommodate interactions among variables, and distinguish patterns occurring at different temporal scopes.

ProtoTS provides a related concept-like approach to interpretable time-series forecasting by learning hierarchical latent prototypes associated with canonical forecast curves~\cite{peng2025protots}. Although these prototypes support inspection, they are unnamed, and forecasts can draw on learned information beyond the displayed prototype matches. Users must therefore infer their semantics after training, while the matches alone do not fully account for a prediction. In contrast, \sys~uses an offline LLM to propose named temporal concepts and executable labeling rules, then organizes the supervised concepts across historical context, local forecast intervals, and the full forecast horizon. This design gives the forecasting pathway a stable semantic vocabulary that supports direct inspection and intervention without manual segment-level annotation.

%% file: method.tex
\section{Method}

\begin{figure*}[!t]
    \centering
    \includegraphics[width=\textwidth]{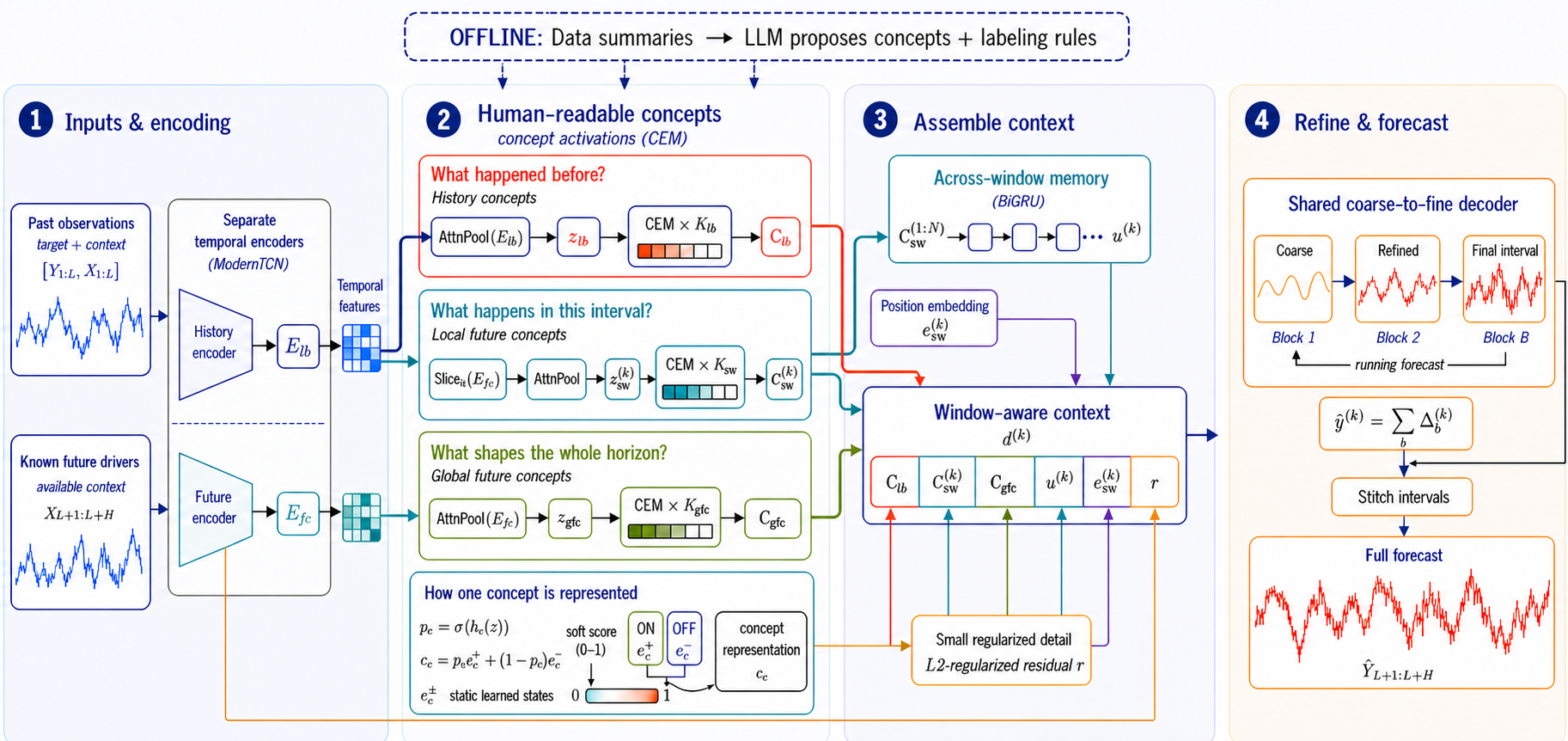}
    \caption{Overall architecture of \sys~for a single forecast sub-window. An LLM proposes three concept sets covering the lookback window, each forecast sub-window, and the global forecast horizon. A ModernTCN-style encoder embeds the input; the temporal slices of the embedding drive the concept activation probabilities, which combine with two static concept embeddings to form concept-context vectors. These are concatenated with a positional embedding, cross-sub-window BiGRU output, and a residual channel, and fed into a stacked residual decoder that refines the forecast in a coarse-to-fine manner.}
    \label{fig:conceptts}
\end{figure*}

\subsection{\textbf{Formulation and Framework}}
\subsubsection{\textbf{Problem Formulation}}

In this project, we consider the multivariate time series forecasting problem in which a group of exogenous features is used as context information to help forecast an endogenous (target) feature. Given a look-back target sequence $\mathbf{Y}_{1:L} \in \mathbb{R}^{L}$, look-back covariates $\mathbf{X}_{1:L} \in \mathbb{R}^{L \times C}$, and
covariates available over the forecast horizon $\mathbf{X}_{L+1:L+H} \in \mathbb{R}^{H \times C}$, our model $\mathcal{F}$ predicts $\widehat{\mathbf{Y}}_{L+1:L+H} = \mathcal{F}(\mathbf{Y}_{1:L}, \mathbf{X}_{1:L}, \mathbf{X}_{L+1:L+H})$. In addition to this future-aware setup, we also evaluate a look-back-only setting that removes access to $\mathbf{X}_{L+1:L+H}$ and forces the model to forecast from $\mathbf{Y}_{1:L}$ and $\mathbf{X}_{1:L}$ alone.

\subsubsection{\textbf{Framework Overview}}

Fig.\ref{fig:conceptts} illustrates the ConceptTS pipeline. An offline LLM proposer generates three concept sets --- for the lookback window, forecast sub-windows, and the full forecast horizon --- each paired with an executable predicate that labels every training segment. A ModernTCN-style encoder preserves per-channel and temporal structure, producing embeddings that drive concept activations in three parallel CEM bottlenecks. A shared residual sub-decoder assembles the concept contexts, cross-sub-window BiGRU output, and a regularized residual channel into the final forecast, trained end-to-end with the hybrid loss.

\subsection{\textbf{LLM-driven Concept Proposer}}

Instead of relying on human-authored concept sets and per-segment labels (the standard CBM/CEM requirement), we provide an offline LLM with dataset statistics and ask it to return three concept sets along with executable Python predicates that label every training segment.

\subsubsection{\textbf{Per-Segment Statistical Summary}}

For each input segment $s_{i}$, we first extract the structured statistical summary from these four aspects:

\begin{itemize}
    \item \textbf{Look-back statistics:} \{mean, std, slope, range\} per
    channel; plus FFT-derived \{dominant period, 24h amplitude, spectral
    entropy, low-frequency ratio\} to expose cyclic patterns.

    \item \textbf{Forecast statistics:} \{mean, std, slope, range\} per
    non-target feature channel over the horizon.

    \item \textbf{Sub-window statistics:} \{mean, std, slope, range\} per
    sub-window for finer-grained descriptions.

    \item \textbf{Cross-channel and time context:} pairwise Pearson
    correlations between key channel pairs; look-back/horizon start-end
    timestamps; season tag (spring/summer/autumn/winter).
\end{itemize}

Combined together, these information gives the LLM a comprehensive view of what each segment looks like, facilitating its following concept proposal and per-concept activation labeling.

\subsubsection{\textbf{Summary Vector Clustering and Profiling}}

To provide the LLM with realistic guidance about the dataset's regime
structure, the summary vectors are clustered into $N_c$ groups via
$K$-means. To prevent leakage, this clustering and all per-cluster and
global statistics referenced below are computed on the training split
only. For each cluster $c$, we compute a compact profile containing
(i) its size and fraction of the training set, (ii) for each continuous
channel $j$, the z-score
$(\mu_{c,j} - \mu_{\mathrm{global},j}) / \sigma_{\mathrm{global},j}$
of the cluster's mean against the global (training) distribution, and
(iii) for each categorical channel $j$, the dominant value and its
purity (fraction of cluster members sharing that value). The $N_c$
profiles are included in the LLM prompt alongside global per-channel
statistics, providing the LLM with data-based local statistics to
facilitate its concept generation without extending the prompt too
long.

\subsubsection{\textbf{LLM Concept Proposal Generation}}

For the initial concept generation, we provide the LLM with a comprehensive prompt containing background information, channel
descriptions, global statistical summaries, cluster profiles, the
segment scheme, the task specification, and the output format. A key
design choice is that we ask the LLM to generate three types of concept
sets, each targeting a different input aspect:
\begin{itemize}
    \item $\textbf{look-back Concepts}$ describe semantic patterns observed in the look-back window, providing history context and general trends information. Examples: ``\textit{is look-back window hot on average?}'', ``\textit{is history rainy in general?}''
    \item $\textbf{Forecast Sub-window Concepts}$ describe fine-grained local details of the context features within the sub-window over the forecasting horizon (e.g., ``\textit{is the sub-window in morning rush period?}'', ``\textit{is the sub-window wind speed rising?}'')
    \item $\textbf{Global Forecast Concepts}$ describe general characteristics of the forecasting horizon (e.g., ``\textit{is forecasting horizon observing extreme temperature change?}'', ``\textit{is forecasting horizon experiencing large pressure swing?}''). Such global forecast concepts serve as a supplement to forecast sub-window concepts, extracting macro patterns over the full forecasting horizon.
\end{itemize}
For both Forecast Sub-window and Global Forecast concepts, the prompt
explicitly instructs the LLM to build the predicates from context
covariates only and never from the target variable itself, since the
target values in the forecast horizon are unavailable to the model at
inference time.

Unlike similar work in other domains that prompt the LLM only for
natural-language descriptions of concepts, we explicitly require the
language model to return an executable Python predicate that evaluates
whether a segment matches the concept, using the precomputed statistics
as input. The LLM proposer itself is invoked only once to generate
these predicates; the concept labels for every training segment are
then produced automatically by executing each predicate over the
segment's precomputed statistics, avoiding costly per-segment LLM
calls.

\subsubsection{\textbf{Concept Filtering}}

After proposing $N_{initial} = N_{look-back} + N_{sub-window} + N_{forecast}$ concepts, we compute each concept's positive rate $r_i$ over the training set and filter out concepts that are not useful by retaining only concepts with $r_i \in [r_{\min}, r_{\max}]$. After filtering, $N_{concept}$ remain.  During the concept evaluating process, the binary labels derived for each segment, $y_{i} \in \{0, 1\} ^ {N_{concept}}$, become the training target of the concept-supervision losses.

\subsection{\textbf{Convolution-based Encoder}}

ProtoTS's feature fusion and temporal mean pooling form compact prototype representations, while our multiscale concepts require variable-specific local evidence, leading us to adopt ModernTCN's variable-independent embedding to preserve channel identity and its large-kernel depthwise convolution to capture broad temporal context \cite{luo2024moderntcn}.

Given an input of shape $(B, T, C)$, each variable channel would be encoded separately and be projected into the same embedding dimension, D, and all embeddings are concatenated along the variable axis to form a tensor of shape $(B, T, C, D)$. Next, a depthwise convolution with stride S is used to patchify the tensor into shape $(B, C, D, N)$, with $N = \frac{T}{S}$. Each ModernTCN block combines a large-kernel depthwise temporal convolution with two grouped pointwise inverted bottlenecks that mix information along the feature axis on a per-variable basis and along the variable axis on a per-feature basis, respectively. Each block is wrapped in a residual connection. A final linear projection over the patch axis maps the $N$ patch positions back to the original $T$ timesteps, yielding the output embedding $\mathbf{E} \in \mathbb{R}^{B \times T \times D_{\text{enc}}}$. We preserve the time axis to provide more comprehensive information to subsequent concept bottleneck module. Overall, the encoder structure produces embedding that is rich in semantic meaning along both feature and temporal axes.

In our default setting, we use two parallel encoders to encode the look-back window $\mathbf{Y}_{1:L}, \mathbf{X}_{1:L}$ and the forecasting horizon $\mathbf{X}_{L+1:L+H}$ separately. We do this to manually set an information barrier between the two time slots.

\subsection{\textbf{Parallel Concept Embedding Bottleneck Module}}

After encoding the input segments, the embedding is then passed into three parallel concept embedding bottlenecks: a look-back concept bottleneck summarizing \textit{what the recent history context looks like}, $N_{\mathrm{sw}}$ forecasting sub-window concept bottlenecks each capturing \textit{what key patterns the other covariates display in this portion of prediction horizon}, and a global forecasting concept bottleneck telling \textit{what macro trends and behaviors the context feature channels expose in the forecasting window}. By feeding in these three concept embeddings into the decoder, the model can make accurate forecasting, and the prediction is highly interpretable by checking the activations of the concepts in each of the three concept bottleneck modules.

\subsubsection{\textbf{Attention-weighted Context Pooling}}

As mentioned in the last section, the embeddings coming out from the encoder are denoted by $\mathbf{E_{look-back}} \in \mathbb{R}^{B \times T_{look-back} \times D_{\text{enc}}}$ and $\mathbf{E_{forecast}} \in \mathbb{R}^{B \times T_{forecast} \times D_{\text{enc}}}$. In order to make three concept bottlenecks summarizing the correct temporal portions, we first use three attention-weighted pooling modules to convert the embeddings into three corresponding context vectors:

$\mathbf{z}_{\text{lb}} = \text{AttentionPool}(\mathbf{E}_{look-back}\bigl[1:T_{look-back}\bigr])$,

$\mathbf{z}_{\text{sw}}^{(k)} = \text{AttentionPool}(
    \mathbf{E}_{forecast}\bigl[(k-1)\,T_{\text{sw}} + 1 \,:\, k\,T_{\text{sw}}\bigr])
$,

$\mathbf{z}_{\text{gfc}} = \text{AttentionPool}(\mathbf{E}_{forecast}\bigl[1 : T_{forecast}\bigr])$

in which $T_{look-back}$, $T_{forecast}$, and $T_{\text{sw}}$ represent the number of time steps in the look-back window, the number of time steps in the forecasting horizon, and the number of time steps in each sub-window respectively.

For any temporal slice $\mathbf{E} \in \mathbb{R}^{B \times T^{'} \times D}$, the $\text{AttentionPool}(\cdot)$ computes an additive-attention score per timestep, softmaxes them along the time axis, and returns the weighted sum:
\begin{equation}
    \boldsymbol{\alpha} = \mathrm{softmax}\!\left(\mathbf{W}_2\, \tanh(\mathbf{W}_1 \mathbf{E})\right),
    \qquad
    \mathbf{z} = \sum_{t=1}^{T^{'}} \alpha_t\, \mathbf{E}_t \in \mathbb{R}^{D},
\end{equation}
where $\mathbf{W}_1$ and $\mathbf{W}_2$ are the two layers of a small
attention MLP.

We use additive attention instead of the time-axis mean pooling adopted in ProtoTS: mean pooling weights every timestep equally and dilutes any temporally-localized signal into a single average, whereas additive attention learns content-dependent weights that emphasize the semantically important timesteps and suppress the rest, giving the concept bottlenecks a sharper temporal summary.

\subsubsection{\textbf{Concept Embedding Module}}
We adapt the full concept embedding model (CEM) structures in our model, which is an alternative option to the original concept bottleneck model (CBM) \cite{espinosa2022cem}. In CEM, the expressive capacity of each concept is enhanced by two high dimensional embeddings, one represents \textit{when the concept is active} ($\mathbf{e}_c^{+}$) and the other one represents \textit{when the concept is absent} ($\mathbf{e}_c^{-}$). One important innovation is we design $\mathbf{e}_c^{+}$ and $\mathbf{e}_c^{-}$ to be \textbf{input-independent}, meaning all concept embeddings are shaped through learning and kept static during inference stage. Such design is to make sure there's no strong information leakage that undermines the strength of interpretability.

A small scoring head $h_c$ takes the encoder output embedding $\mathbf{z}$
and produces a scalar logit, which gives the concept activation
probability after a sigmoid activation function:
\begin{equation}
    p_c(\mathbf{z}) = \sigma\!\left(h_c(\mathbf{z})\right).
\end{equation}

The final concept context vector, for each concept, is the corresponding weighted mixture of the two concept embeddings:
\begin{equation}
    \mathbf{c}_c(\mathbf{z}) = p_c(\mathbf{z}) \cdot \mathbf{e}_c^{+}
    + \left(1 - p_c(\mathbf{z})\right) \cdot \mathbf{e}_c^{-}
    \in \mathbb{R}^{d_c}.
\end{equation}
Stacking across all $N$ concepts within the bottleneck yields the bottleneck
output:
\begin{equation}
    \mathbf{C}(\mathbf{z}) = \left[\mathbf{c}_1(\mathbf{z});\, \mathbf{c}_2(\mathbf{z});\, \ldots;\, \mathbf{c}_N(\mathbf{z})\right]
    \in \mathbb{R}^{N \cdot d_c}.
\end{equation}

\subsubsection{\textbf{Sequence-aware Sub-window Connection}}

Since forecast sub-windows are temporally correlated (e.g., the effect of
high wind speed depends on the humidity of the preceding sub-window), we
make the decoder sub-window-aware. Each sub-window's concept context
$\mathbf{C}_{\text{sw}}^{(k)}$ is projected to a summary vector
$\mathbf{s}^{(k)} = \mathbf{W}_{\text{summ}}\,\mathbf{C}_{\text{sw}}^{(k)}$
and passed through a bidirectional GRU:

\begin{equation}
    \left[\mathbf{u}^{(1)}, \ldots, \mathbf{u}^{(N_{\text{sw}})}\right]
    = \mathrm{BiGRU}\!\left(\left[\mathbf{s}^{(1)}, \ldots,
    \mathbf{s}^{(N_{\text{sw}})}\right]\right).
\end{equation}

The output $\mathbf{u}^{(k)}$ is concatenated into the sub-decoder's input
at sub-window $k$, providing cross-sub-window context. Using the concept
bottleneck output $\mathbf{C}_{\text{sw}}^{(k)}$ (rather than the encoder
embedding) as the BiGRU input prevents information from leaking around the
bottleneck.

\begin{table*}[!t]
    \centering
    \caption{Forecasting performance in the \emph{future-aware} setting
    across three Beijing monitoring stations. Values are mean $\pm$ std
    in $\mu$g/m$^3$ across three training seeds. Cells shaded
    \colorbox{lossbg}{light red} mark a baseline that outperforms
    ConceptTS by more than $10\%$; cells shaded \colorbox{winbg}{light
    green} mark a baseline that underperforms ConceptTS by more than
    $10\%$. Unshaded cells fall within $\pm 10\%$ of ConceptTS. Lower
    is better.}
    \label{tab:future-aware}
    \renewcommand{\arraystretch}{1.15}
    \scriptsize
    \setlength{\tabcolsep}{3pt}
    \begin{tabular}{clcccccccc}
        \toprule
        Station & Metric & \sys & Informer & TFT & Random Forest & XGBoost & LightGBM & DeepAR & NHiT \\
        \midrule
        \multirow{2}{*}{\textbf{Aotizhongxin}}
            & MAE  & $12.71 \pm 0.42$ & \cellcolor{winbg}$28.62 \pm 1.81$ & $12.44 \pm 0.33$ & \cellcolor{winbg}$26.73 \pm 0.02$ & $12.34 \pm 0.01$ & $12.46 \pm 0.01$ & $13.97 \pm 0.70$ & \cellcolor{winbg}$16.94 \pm 0.86$ \\
            & RMSE & $20.61 \pm 0.61$ & \cellcolor{winbg}$47.58 \pm 3.44$ & $20.92 \pm 0.44$ & \cellcolor{winbg}$42.22 \pm 0.03$ & $19.19 \pm 2.79$ & $20.55 \pm 0.03$ & $21.99 \pm 0.32$ & \cellcolor{winbg}$24.80 \pm 1.42$ \\
        \midrule
        \multirow{2}{*}{\textbf{Dingling}}
            & MAE  & $12.25 \pm 0.26$ & \cellcolor{winbg}$24.87 \pm 0.45$ & $12.45 \pm 1.16$ & \cellcolor{winbg}$23.31 \pm 1.23$ & $11.48 \pm 0.01$ & $11.52 \pm 0.01$ & $11.86 \pm 0.53$ & \cellcolor{winbg}$17.03 \pm 0.47$ \\
            & RMSE & $27.44 \pm 0.31$ & \cellcolor{winbg}$42.58 \pm 0.90$ & $27.78 \pm 0.55$ & \cellcolor{winbg}$39.20 \pm 1.40$ & $27.57 \pm 0.02$ & $27.37 \pm 0.02$ & $26.58 \pm 0.68$ & \cellcolor{winbg}$30.92 \pm 0.78$ \\
        \midrule
        \multirow{2}{*}{\textbf{Tiantan}}
            & MAE  & $13.62 \pm 0.68$ & \cellcolor{winbg}$26.06 \pm 0.80$ & $14.35 \pm 1.38$ & \cellcolor{winbg}$28.34 \pm 1.41$ & \cellcolor{lossbg}$12.00 \pm 0.01$ & $12.29 \pm 0.03$ & $13.01 \pm 0.31$ & \cellcolor{winbg}$16.13 \pm 0.19$ \\
            & RMSE & $23.76 \pm 0.82$ & \cellcolor{winbg}$42.92 \pm 1.89$ & $23.64 \pm 2.67$ & \cellcolor{winbg}$43.88 \pm 2.18$ & \cellcolor{lossbg}$20.05 \pm 0.05$ & \cellcolor{lossbg}$20.07 \pm 0.02$ & $21.72 \pm 0.25$ & $24.71 \pm 0.61$ \\
        \bottomrule
    \end{tabular}
\end{table*}

\begin{table*}[!t]
    \centering
    \caption{Forecasting performance in the \emph{look-back-only}
    setting. Cell are colored the same way as TABLE~\ref{tab:future-aware}}
    \label{tab:look-back-only}
    \renewcommand{\arraystretch}{1.15}
    \footnotesize
    \setlength{\tabcolsep}{4pt}
    \begin{tabular}{clccccccc}
        \toprule
        Station & Metric & \sys & TimeXer & iTransformer & Informer & Crossformer & DLinear & LSTM \\
        \midrule
        \multirow{2}{*}{\textbf{Aotizhongxin}}
            & MAE  ($\mu\mathrm{g}/\mathrm{m}^3$) & $55.59 \pm 1.42$ & $57.43 \pm 0.33$ & $59.16 \pm 0.50$ & $55.04 \pm 0.82$ & $54.78 \pm 0.95$ & $56.27 \pm 0.09$ & $55.91 \pm 1.78$ \\
            & RMSE ($\mu\mathrm{g}/\mathrm{m}^3$) & $86.05 \pm 1.56$ & $86.12 \pm 0.55$ & $89.16 \pm 1.26$ & $86.67 \pm 0.61$ & $81.95 \pm 1.38$ & $79.13 \pm 0.05$ & $85.99 \pm 3.02$ \\
        \midrule
        \multirow{2}{*}{\textbf{Dingling}}
            & MAE  ($\mu\mathrm{g}/\mathrm{m}^3$) & $47.23 \pm 0.51$ & $47.84 \pm 0.17$ & $49.28 \pm 0.57$ & $45.22 \pm 0.64$ & $44.84 \pm 0.17$ & $47.11 \pm 0.03$ & $45.52 \pm 1.20$ \\
            & RMSE ($\mu\mathrm{g}/\mathrm{m}^3$) & $73.32 \pm 0.68$ & $71.33 \pm 0.47$ & $72.95 \pm 0.37$ & $69.59 \pm 2.65$ & $68.81 \pm 2.20$ & \cellcolor{lossbg}$65.43 \pm 0.02$ & $68.82 \pm 1.74$ \\
        \midrule
        \multirow{2}{*}{\textbf{Tiantan}}
            & MAE  ($\mu\mathrm{g}/\mathrm{m}^3$) & $57.22 \pm 0.15$ & $58.61 \pm 0.77$ & $59.30 \pm 0.62$ & $54.54 \pm 0.05$ & $55.12 \pm 0.24$ & $55.81 \pm 0.14$ & $54.63 \pm 0.24$ \\
            & RMSE ($\mu\mathrm{g}/\mathrm{m}^3$) & $86.34 \pm 1.47$ & $88.52 \pm 0.68$ & $89.09 \pm 0.78$ & $84.75 \pm 0.60$ & $83.85 \pm 1.13$ & $79.50 \pm 0.16$ & $83.61 \pm 1.09$ \\
        \bottomrule
    \end{tabular}
\end{table*}

\subsubsection{\textbf{Residual Sub-decoder Stack}}
Eventually, the forecast is produced by a shared sub-decoder that takes in a concatenation of the concept bottleneck output embeddings of all three types, the cross-sub-window context embedding, a positional sub-window id embedding, and a narrow residual bypass:
\begin{equation}
    \mathbf{d}^{(k)} = \left[\mathbf{C}_{\mathrm{lb}} \,;\, \mathbf{C}_{\mathrm{sw}}^{(k)} \,;\, \mathbf{C}_{\mathrm{gfc}} \, ; \mathbf{u}^{(k)} \,;\, \mathbf{e}_{\mathrm{sw}}^{(k)} \,;\, \mathbf{r}_{\mathrm{lb}} \,;\, \mathbf{r}_{\mathrm{sw}}^{(k)}\right],
\end{equation}
with $\mathbf{e}_{\mathrm{sw}}^{(k)} = \mathrm{Embedding}(N_{\mathrm{sw}}, d_{\mathrm{id}})(k)$, a positional-only embedding that carries no data-derived signal. $\mathbf{C}_{\mathrm{lb}}$ and $\mathbf{C}_{\mathrm{gfc}}$ represent the eventual output from the look-back concept bottleneck and the global forecast concept bottleneck, respectively. The last two terms form an optional residual bypass, with $\mathbf{r}_{\mathrm{lb}} = \mathbf{W}_{\mathrm{lb}}\mathbf{z}_{\mathrm{lb}}$ shared across all sub-windows and $\mathbf{r}_{\mathrm{sw}}^{(k)} = \mathbf{W}_{\mathrm{sw}}\mathbf{z}_{\mathrm{sw}}^{(k)}$ computed per sub-window, both linear projections of the pooled look-back and forecast sub-window encoder features into a low-dimensional space $\mathbb{R}^{r}$. This bypass carries the continuous information that the soft scalar
concept activations cannot fully express. It is kept low-dimensional
and regularized by a dedicated loss term to discourage the model from
relying too much on the residual connection; the weight
$\alpha_{\mathrm{res}}$ of this regularization is a direct knob for
trading off forecast accuracy against how much of the prediction is
routed through the concept bottlenecks.

Inspired by NHiTS, the decoder is a total of $N_{dec}$ stacked MLP blocks, with each block refining the output taken from its previous block \cite{challu2023nhits}. 

\begin{equation}
    \hat{\mathbf{y}}^{(k)} = \sum_{b=1}^{B_{\mathrm{dec}}} \boldsymbol{\Delta}_b^{(k)},
    \qquad
    \boldsymbol{\Delta}_b^{(k)} = \mathrm{Block}_b\!\left(\mathbf{i}_b^{(k)}\right).
\end{equation}
We also use decoder feedback structure, meaning block $b > 1$
additionally sees the running forecast so far:
$\mathbf{i}_1^{(k)} = \mathbf{d}^{(k)}$ and
$\mathbf{i}_b^{(k)} = \left[\mathbf{d}^{(k)} \,;\, \mathrm{flatten}\!\left(\sum_{b' < b} \boldsymbol{\Delta}_{b'}^{(k)}\right)\right]$
for $b > 1$. Essentially, the stacked sub-decoder module is constantly using the information summarized by the three types of concept bottlenecks along with the neighboring sub-windows' contexts to improve its forecasting, from initial coarse forecasting to eventual high-quality prediction.

The shared decoder outputs forecasting $\hat{\mathbf{y}}^{(k)} \in \mathbb{R}^{B \times T_{\mathrm{sw}} \times 1}$ for the $k$-th sub-window. Then, all partial predictions are concatenated along time axis to form the final forecasting curve.

\subsection{\textbf{Training Pipeline}}

The two parallel encoders, the three CEM modules, and the decoder are
trained jointly end-to-end. We use a hybrid loss function to make sure
different parts of the model work as expected:
\begin{equation}
\begin{split}
    \mathcal{L} = {} & \alpha_{\mathrm{f}}\, \mathcal{L}_{\mathrm{forecast}}
    + \alpha_{\mathrm{lb}}\, \mathcal{L}_{\mathrm{concept}}^{\mathrm{lb}}
    + \alpha_{\mathrm{fc}}\, \mathcal{L}_{\mathrm{concept}}^{\mathrm{fc}} \\
    & + \alpha_{\mathrm{gfc}}\, \mathcal{L}_{\mathrm{concept}}^{\mathrm{gfc}}
    + \alpha_{\mathrm{bd}}\, \mathcal{L}_{\mathrm{boundary}} \\
    & + \alpha_{\mathrm{cc}}\, \mathcal{L}_{\mathrm{collapse}}
    + \alpha_{\mathrm{residual}}\, \mathcal{L}_{\mathrm{residual}},
\end{split}
\end{equation}

The full objective combines seven weighted loss terms: (i) an $L_1$ forecast loss on the target; (ii)–(iv) three BCE concept losses on the LLM-derived binary labels for the lookback, forecast sub-window, and global forecast bottlenecks, forcing the predicted activation score of each concept to be close to the LLM-derived labels; (v) an $L_1$ boundary-smoothness penalty across sub-window boundaries, preventing discontinuous values in the final forecast output; (vi) a variance floor that keeps each concept's activation from
collapsing to a near-constant (and therefore uninformative) value
across inputs; and (vii) an $L_2$ regularizer on the residual channel that directly controls the accuracy-interpretability trade-off.

%% file: experiments.tex
\section{Experiments}

\subsection{\textbf{Model Performance}}

\subsubsection{\textbf{Dataset}}

We choose to use Beijing Multi-Site Air Quality dataset in this project, which is an open-source multivariate time series dataset provided by UC Irvine Machine Learning Repository \cite{chen2019beijingair}. This dataset contains 6 pollutants concentrations and 5 relevant meteorological context features at 12 air quality monitoring sites in Beijing, and the time spans from March 1st, 2013 to February 28th, 2017, at a one-hour recording interval. We choose to use PM2.5 concentration as our target feature, and the context features include five other pollutants (PM10, SO2, NO2, CO, O3) and five meteorological measurements (temperature, pressure, dew point, precipitation, wind speed). The missing data in each feature channel are filled using linear interpolation.

We benchmark overall forecasting performance on three monitoring sites
--- \textit{Aotizhongxin}, \textit{Dingling}, and \textit{Tiantan} ---
and use \textit{Aotizhongxin} as the representative site for all
subsequent ablation and case-study experiments.

\subsubsection{\textbf{Evaluation Metrics}}

We report the model performance using two standard metrics in time series forecasting: mean absolute error (MAE) and root mean squared error (RMSE).

The two metrics are complementary: MAE shows on average how much the model's prediction deviates from the ground-truth value at each time step, while RMSE punishes large errors more heavily, displaying how well the model can handle the challenging portions of the forecasting like the air pollutant spikes. The metrics are reported in raw $\mu\mathrm{g}/\mathrm{m}^3$, computed by inverting the training-set z-scoring.

\subsubsection{\textbf{Experiment Settings}}

We implement two experiment settings to reflect two natural data availability regimes that are both common in multivariate time series forecasting domain. In the \textbf{future-aware} setting (the default setting clarified in section 3.1.1), the exogenous features are assumed to be known beforehand for the forecasting horizon, which is realistic when the context features come from an upstream prediction system. This is often the case in forecasting tasks related to meteorological factors like wind speed, precipitation, and temperatures. In the \textbf{look-back-only} setting, the model only has accessibility to all features in the history $\{\mathbf{Y}_{1:L}, \mathbf{X}_{1:L}\}$, matching the regular setting that Time-Series-Library uses for its benchmark models.

The two settings share the same encoder-pool-CEM-decoder model structure, but the concept bottleneck types are different. In future-aware setting, the LLM proposer proposes three concept sets for look-back window, forecasting sub-window, and global forecasting horizon respectively. In the look-back-only setting, there is only one encoder that encodes the look-back window (since there is no future data to encode), and the proposer instead proposes two concept sets, one for global look-back window, and one for look-back sub-window since now the model additionally partitions the look-back window into $N_\text{lbsw}$ sub-windows to provide more interpretability and higher prediction accuracy.

Because most Time-Series-Library baselines only consume the look-back window, we compare against a different baseline set in each setting.

\subsubsection{\textbf{Model Evaluation}}

Under future-aware setting, as shown in Table \ref{tab:future-aware}, our model is competitive with state-of-the-art benchmark time forecasters. Taking \textit{Aotizhongxin} as example, our model achieves a $12.71 \mu\mathrm{g}/\mathrm{m}^3$ MAE, outperforming Informer ($28.62 \mu\mathrm{g}/\mathrm{m}^3$) \cite{haoyietal-informer-2021}, Random Forest ($26.73 \mu\mathrm{g}/\mathrm{m}^3$) \cite{breiman2001rf}, DeepAR ($13.97 \mu\mathrm{g}/\mathrm{m}^3$) \cite{salinas2020deepar}, and NHiT ($16.94 \mu\mathrm{g}/\mathrm{m}^3$) \cite{challu2023nhits}, and matches performances of XGBoost ($12.34 \mu\mathrm{g}/\mathrm{m}^3$) \cite{chen2016xgboost}, TFT ($12.44 \mu\mathrm{g}/\mathrm{m}^3$) \cite{lim2020temporalfusiontransformersinterpretable}, and LightGBM ($12.46 \mu\mathrm{g}/\mathrm{m}^3$) \cite{ke2017lightgbm}, while providing concept-level explanations these models cannot offer. Consequently, our model contributes to the multivariate time series forecasting by achieving SOTA accuracy while directly incorporating human-like concept-based decision making process into the forecasting pipeline.

For the look-back-only setting, we compare our model with several Time-Series-Library benchmarks and the results are shown in Table \ref{tab:look-back-only}. The results illustrate that our model's prediction accuracy is also competitive with SOTA time series forecasters even only taking in historical context. For the station \textit{Aotizhongxin}, \sys~($55.59 \mu\mathrm{g}/\mathrm{m}^3$ MAE) beats TimeXer ($57.43 \mu\mathrm{g}/\mathrm{m}^3$) \cite{wang2024timexer}, iTransformer ($59.16 \mu\mathrm{g}/\mathrm{m}^3$) \cite{liu2023itransformer}, DLinear ($56.27 \mu\mathrm{g}/\mathrm{m}^3$) \cite{Zeng2022AreTE}, and LSTM ($55.91 \mu\mathrm{g}/\mathrm{m}^3$) \cite{hochreiter1997lstm}. Notably, although the other two models (Informer, Crossformer) yield lower MAE, our model remains highly competitive, underperforming them by only $0.55 \mu\mathrm{g}/\mathrm{m}^3$ and $0.81 \mu\mathrm{g}/\mathrm{m}^3$ respectively \cite{haoyietal-informer-2021} \cite{zhang2023crossformer}.

In both cases, the gaps on RMSE are somewhat larger, suggesting \sys~handles typical timesteps well but is slightly less sensitive to extreme pollution spikes, a limitation we leave to future work.

\subsection{\textbf{Ablation Study}}

We conduct experiments to study how different structure designs contribute to the performance of the model. Specifically, we study (i) how the number of forecast sub-window concepts affect forecast accuracy, and (ii) how the residual-channel weight $\alpha_{residual}$ trades off accuracy against interpretability.

\subsubsection{\textbf{Number of Concepts}}

To study the influence of the number of concepts on forecasting accuracy, we fix both the look-back and global forecast concept counts at 10. This gives the model enough baseline capacity to forecast accurately without letting it depend too heavily on these two bottlenecks. Then, we sweep the forecast sub-window concept number to see how the MAE on the test set changes. In this experiment, we disable the residual connection so all forecasting must go through the concept bottleneck entirely.

\begin{figure}[!t]
    \centering
    \includegraphics[width=\linewidth]{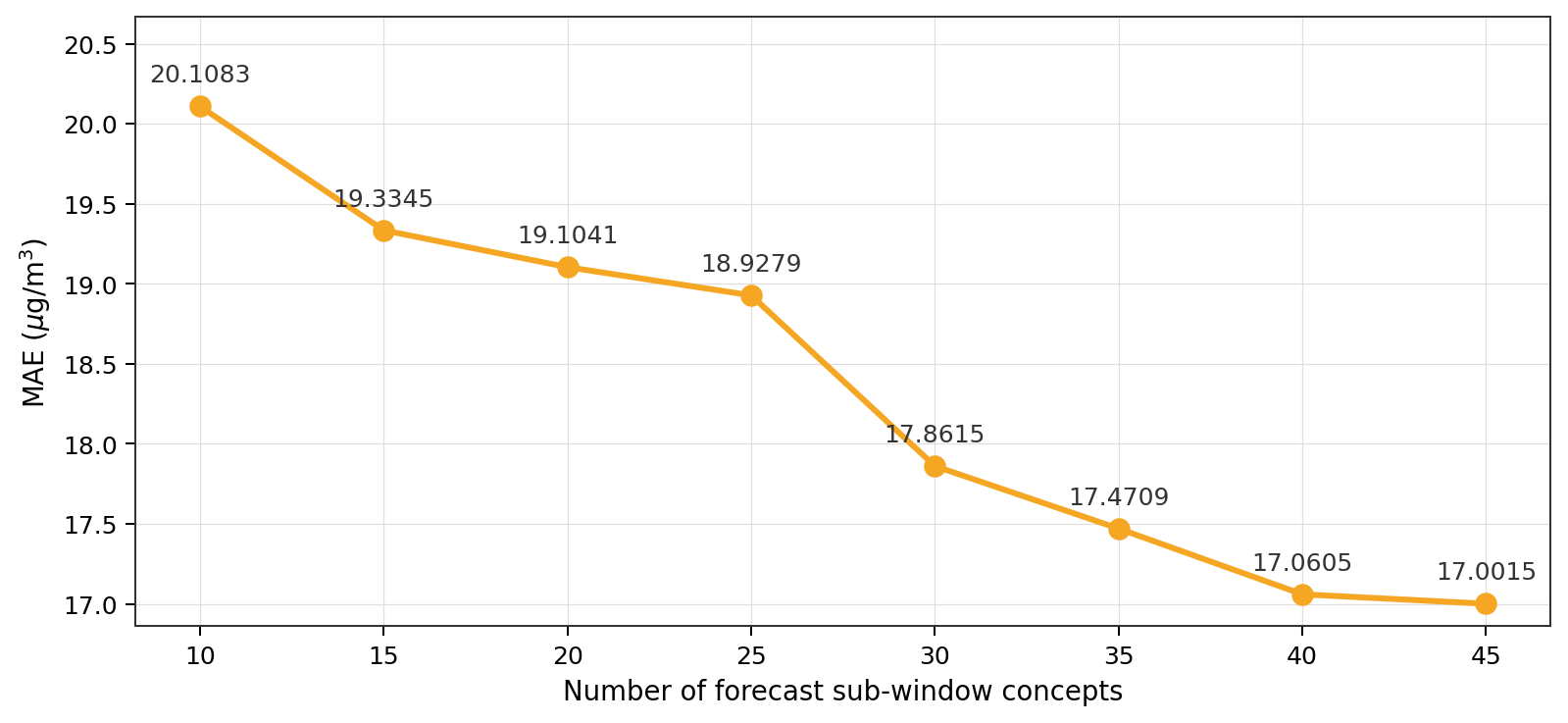}
    \caption{Effect of the number of forecast sub-window concepts on forecast MAE. The $x$-axis reports the number of forecast sub-window concepts(with the number of look-back and global forecast concepts fixed at $10$ each).}
    \label{fig:mae-vs-nfc}
\end{figure}

For each sub-window concept count $N_\text{sw}$, we run three LLM proposals with different random seeds and train the model separately for each; the reported test MAE is the average over these three runs. Such design mitigates the influence of individual LLM-proposal quality and lets us observe the general trend. As shown in Fig \ref{fig:mae-vs-nfc}, the results indicate that MAE decreases monotonically as $N_\text{sw}$ increases. This confirms that the model genuinely relies on combinations of concepts to forecast, and it does hurt the model's performance significantly when the available concepts are limited.

\subsubsection{\textbf{Trade-off between Accuracy and Interpretability}}

In Method section, we introduced the residual channel and its weight $\alpha_{residual}$; this section studies empirically how $\alpha_{residual}$ balances forecast accuracy and interpretability.

We use the same group of concept sets proposed by LLM to make the comparison fair, then we sweep the $\alpha_{\mathrm{residual}}$ from 0.01 to 1.0. Table~\ref{tab:alpha-residual} reports the results. A larger $\alpha_{\mathrm{residual}}$ penalizes the residual connection more strongly, forcing the decoder to rely more on the concept bottleneck. As $\alpha_{\mathrm{residual}}$ decreases, the full-model MAE improves ($15.35 \rightarrow 12.71$), but the gap between full-model and concept-only MAE --- the ``MAE improvement'' column --- widens from $0.5\%$ to $161.0\%$, indicating that the decoder increasingly bypasses the bottleneck via the residual path. $\alpha_{\mathrm{residual}}$ thus provides a direct knob for trading accuracy against interpretability.

\begin{table}[!t]
    \centering
    \caption{Ablation on the residual-channel weight
    $\alpha_{\mathrm{residual}}$. MAE values are in $\mu$g/m$^3$; all
    values are mean $\pm$ std across three training seeds. Improvement is
    the relative gap between concept-only and full-model MAE; larger
    values indicate the decoder relies more on the residual channel.}
    \label{tab:alpha-residual}
    \renewcommand{\arraystretch}{1.2}
    \footnotesize
    \setlength{\tabcolsep}{4pt}
    \begin{tabular}{cccc}
        \hline
        $\alpha_{\mathrm{residual}}$ & Full-model MAE & Concept-only MAE & Improvement \\
        \hline
        0.01 & $12.71 \pm 0.42$ & $33.07 \pm 5.19$ & $\uparrow 161.0\% \pm 47.6\%$ \\
        0.1  & $13.65 \pm 0.64$ & $18.75 \pm 3.90$ & $\uparrow 38.3\% \pm 34.8\%$  \\
        1.0  & $15.35 \pm 0.54$ & $15.43 \pm 0.59$ & $\uparrow 0.5\% \pm 0.3\%$    \\
        \hline
    \end{tabular}
\end{table}

\begin{figure*}[!t]
    \centering
    \includegraphics[width=\textwidth]{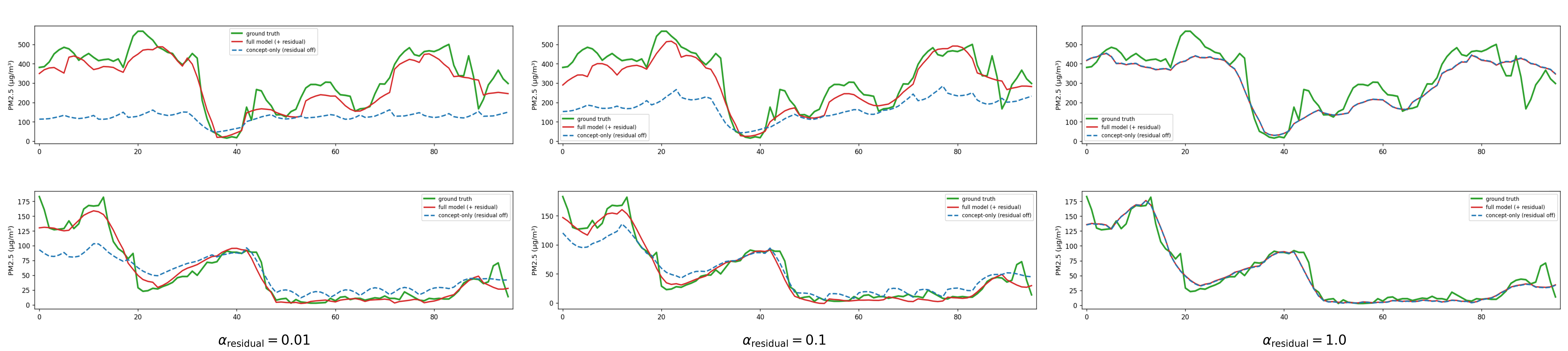}
    \caption{Comparison of full model forecasting versus concept-only forecasting under different $\alpha_{\mathrm{residual}}$. Each column corresponds to one value of $\alpha_{\mathrm{residual}}$.}
    \label{fig:residual-case-study}
\end{figure*}

The forecasting curves on the test set are shown in
Figure~\ref{fig:residual-case-study}. The two rows display two randomly chosen test windows, and the three columns show the forecasting results at $\alpha_{\mathrm{residual}} = 0.01,\ 0.1,\ \text{and}\ 1.0$. Each plot contains three curves: the ground-truth values, the full-model prediction, and the concept-only prediction. As can be seen from the figure, as $\alpha_{\mathrm{residual}}$ increases, the concept-only curve gradually converges to the full-model curve, indicating that the decoder increasingly bases its prediction on the concept bottlenecks. Another point worth noticing is that, despite the relatively large MAE of the concept-only prediction when $\alpha_{\mathrm{residual}}$ is very small (e.g., $\alpha_{\mathrm{residual}} = 0.01$), the concept-only curve still
captures the correct shape of the ground truth at both the macro and
micro scales. This shows that even when $\alpha_{\mathrm{residual}}$ is small and the model relies heavily on the residual connection for
accurate forecasting, the concept bottlenecks can still provide
meaningful interpretability.

\subsection{\textbf{Case Study}}

\begin{figure*}[!t]
    \centering
    \includegraphics[width=\textwidth]{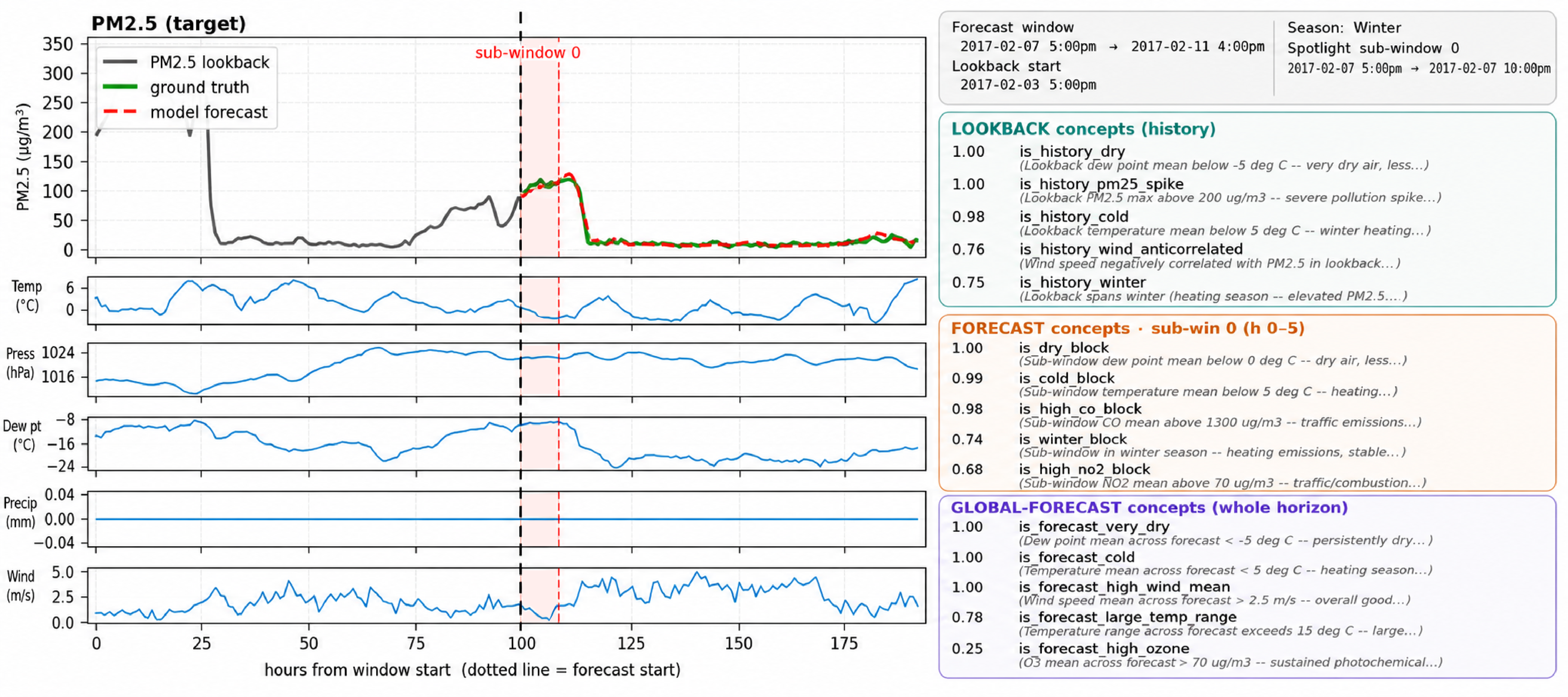}
    \caption{Prediction result of our model on a randomly chosen test
    window, together with the top-5 most activated concepts in each of
    the three concept sets. The black dotted line marks the boundary
    between the look-back window and the forecast horizon; the red
    shaded region indicates the sub-window randomly spotlighted for the
    case study.}
    \label{fig: case_study}
\end{figure*}

For the case studies below we use 20/30/20 concepts for the lookback/forecast sub-window/forecast-global bottlenecks and disable the residual channel to inspect the concept pathway in isolation.

\subsubsection{\textbf{Concepts Activation}}

The forecast for one of the test windows is shown in
Fig.~\ref{fig: case_study}, together with the meteorological context
features (other pollutants are omitted due to space constraints). On the right, we list the top-5 most activated concepts for each of the three concept sets; the gray text in parentheses under each concept is the natural-language description generated by the LLM proposer.

From the figure and the concept lists, we can see that the model
correctly activates the concepts that describe the input. To be more specific, the chosen sample is a cold window in early February, with air temperature oscillating around $0\,^\circ\mathrm{C}$ and dew point constantly below $-8\,^\circ\mathrm{C}$. These features are correctly captured across all three bottlenecks: each of them assigns very high confidence to concepts describing the window as ``cold'' and ``dry'', and the forecast sub-window bottleneck additionally identifies the sub-window
as a ``winter block''. More interestingly, the look-back bottleneck also fires on a concept named ``is\_history\_wind\_anticorrelated'', which describes the pattern where PM2.5 is negatively correlated with
wind speed. This pattern is clearly visible in the look-back window:
the wind speed stays near zero for the first 25 hours while PM2.5 is
high, and as the wind speed rises over the next 50 hours, the PM2.5 level drops sharply to near zero. This ability to expose a learned cross-channel relationship in a human-readable way is exactly what previous black-box models cannot offer --- they may implicitly rely on the same wind--PM2.5 anticorrelation to make their predictions, but they cannot tell the user that they are doing so.

\subsubsection{\textbf{Concept Sweep Study}}

\begin{figure*}[!h]
    \centering
    \includegraphics[width=\textwidth]{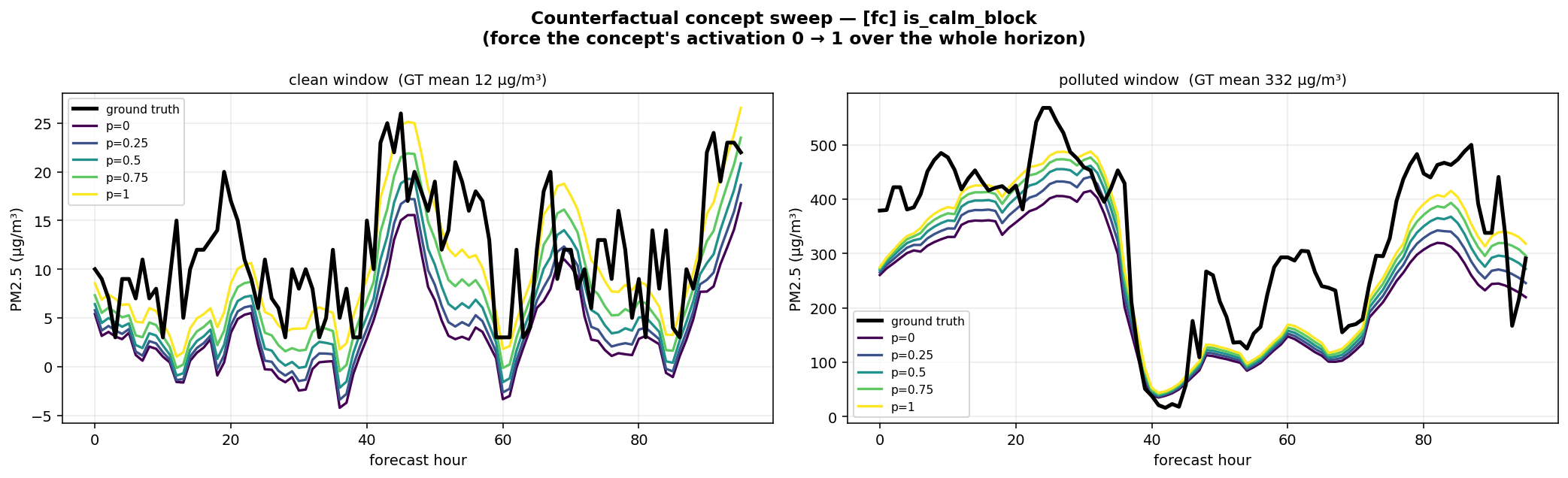}
    \caption{Forecast curves generated when the activation probability
    of a single concept is manually swept from $0$ to $1$. We choose two
    test windows to illustrate: one with a low overall pollutant level
    (left) and one with a high pollutant level (right).}
    \label{fig: sweep}
\end{figure*}

To further explore the model's interpretability, we conduct a counterfactual concept sweep experiment: we manually sweep the activation probability of one concept from $0$ to $1$, while leaving the other concepts' predicted probabilities untouched. This enables us to directly observe how turning one concept on and off influences the model's prediction.

One such example is shown in Fig.~\ref{fig: sweep}. The chosen concept
is the sub-window concept ``is\_calm\_block'', which describes whether
the wind speed stays consistently low during the sub-window. We show
two contrasting scenarios --- a low-pollutant window (ground truth
mean $\approx 12\,\mu\mathrm{g}/\mathrm{m}^3$) and a high-pollutant
window (ground truth mean $\approx 332\,\mu\mathrm{g}/\mathrm{m}^3$).
In both scenarios the concept is positively related to the forecast:
as the activation probability of ``is\_calm\_block'' increases from $0$ to $1$, the overall predicted PM2.5 level rises accordingly. This matches physical intuition, since calm air prevents pollutant
particles from dispersing and therefore leads to a higher local
pollution concentration.

Together, these two case studies illustrate two complementary interpretability properties of our model: the concept activations faithfully describe the input, and intervening on a concept's activation can help a domain expert intuitively understand how the concept contributes to the forecast.

\subsubsection{\textbf{Concept Intervention Diagnostic}}

To further verify that our model genuinely relies on the concept
activation channel to make forecasts, we conduct a concept intervention
diagnostic on the test set. At inference time, we hold the model's
weights unchanged but replace every concept's predicted activation
probability $p_c$ with one of four regimes:
\begin{itemize}
    \item \textbf{baseline}: the model's own predicted activation
    probabilities (no change).
    \item \textbf{oracle}: each concept activation is replaced with its
    ground-truth binary label.
    \item \textbf{flip}: each concept activation is replaced with
    $1 - $ ground-truth label (purely inverse the original prediction).
    \item \textbf{random}: each concept activation is replaced with a
    sample from $\mathrm{Bernoulli}(0.5)$ (``random guess'').
\end{itemize}
The idea is straightforward. If the decoder relies mainly on the concept
activation channel to make forecasts, then flipping the activations to
their opposites should dramatically hurt the forecast quality. On the
other hand, if the decoder mostly bypasses the concepts and reads
information off the CEM's $\pm$ embeddings, then flipping the activations
should have little effect on the forecast.

\begin{table}[!t]
    \centering
    \caption{Concept intervention diagnostic on the test set.}
    \label{tab:concept-intervention}
    \renewcommand{\arraystretch}{1.2}
    \begin{tabular}{lccc}
        \hline
        Regime  & MAE ($\mu$g/m$^3$) & $\Delta$ vs.\ baseline ($\mu$g/m$^3$) \\
        \hline
        baseline  & 15.371 & --- \\
        oracle    & 23.331 & $\downarrow$ 7.960 \\
        flip      & 99.661 & $\downarrow$ 84.290 \\
        random    & 60.818 & $\downarrow$ 45.446 \\
        \hline
    \end{tabular}
\end{table}

The results are shown in Table \ref{tab:concept-intervention}. The
\textbf{flip} intervention increases the test MAE from $15.371$ to
$99.661~\mu\mathrm{g}/\mathrm{m}^3$ (a degradation of $84.990~\mu\mathrm{g}/\mathrm{m}^3$),
and the \textbf{random} intervention increases it to
$60.818~\mu\mathrm{g}/\mathrm{m}^3$ ($\downarrow45.446~\mu\mathrm{g}/\mathrm{m}^3$).
Both degradations are significant. This sharp separation confirms that the concept activation channel carries the majority of the predictive signal --- flipping the concepts fundamentally changes the forecast, which would not happen if the decoder were bypassing the bottleneck. The slight degradation brought by providing the model with real ground-truth concept labels (\textbf{oracle}) is also noticeable. It demonstrates that the soft activation probability assignment learned by the model carries more information than the hard binary category. The model learns to use continuous probability assignment to represent different degree of a concept (``very cold'', ``somewhat cold'', ``slightly cold'').

\subsubsection{\textbf{Failure Mode}}

A concept-level AUROC analysis reveals that shape-dependent concepts --- those describing temporal trends like monotonic rises or falls --- are consistently the hardest for the model to learn, a limitation we trace to the statistics-based label pipeline. Additionally, the framework is better suited to tasks whose target is driven mainly by context features than to those dominated by strong long-term cycles.


%% file: conclusion.tex
\section{Conclusion}

We introduced \sys, an interpretable multivariate time-series forecaster that adapts concept bottlenecks from image classification. An offline LLM constructs named temporal concepts and executable labeling rules, reducing manual annotation and requiring no LLM at inference time. Across three air-quality sites and two settings, \sys~achieves competitive accuracy while supporting concept-level inspection and intervention. Inverting ground-truth concept labels increased MAE about ninefold, showing strong reliance on the concept pathway.

%% file: appendix.tex
\appendices

\section{ProtoTS Failure Mode}

\begin{figure*}[!t]
    \centering
    \includegraphics[width=1.5\columnwidth]{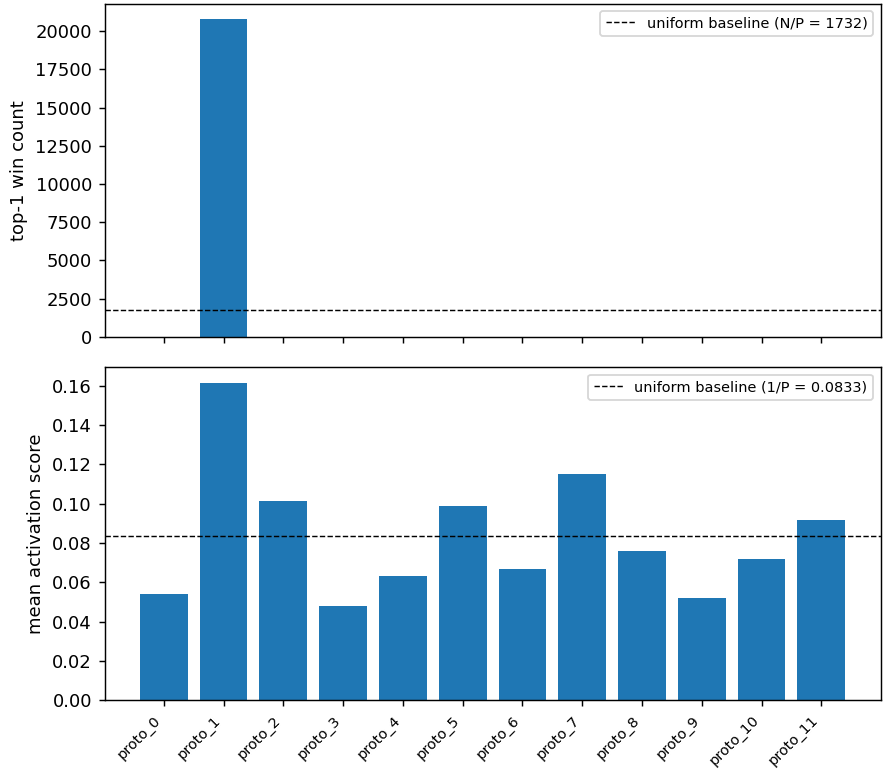}
    \caption{Top-1 win counts (top) and mean activation scores (bottom) of the 12 root prototypes under ProtoTS's default entropy-regularization weight. \texttt{proto\_1} wins the argmax on essentially every test segment, yet the mean activation scores across all prototypes remain tightly clustered around the uniform baseline $1/P = 0.0833$. This gap between argmax dominance and near-uniform activations is the hallmark of the \texttt{value\_bias} bypass: the softmax activations carry almost no per-input information, and the actual forecasting signal flows through the \texttt{value\_bias} $\times$ \texttt{prototype\_value} channel that reads directly from the encoder output.}
    \label{fig:protots_activation}
\end{figure*}

\subsection{ProtoTS Structure}

ProtoTS is a two-stage, hierarchically-prototypical time-series
forecaster that takes the target channel plus a set of context
channels as input.

\textbf{Stage 1: Encoder pretraining.}
Each input channel is first embedded independently --- continuous
features (e.g., skin temperature) through an MLP, and discrete
features (e.g., hour-of-day) through a lookup table. The per-channel embeddings are then passed through a stack of residual MLP-mixer blocks that mix information along both the time and feature axes. In this stage, the encoder is trained to produce both the embedding and the forecast curve, forcing it to learn representations that already carry the information needed for the downstream task.

\textbf{Stage 2: Prototype layer.}
A hierarchical prototype layer is trained jointly with the pretrained encoder. Each prototype consists of two randomly initialized parts: a \emph{prototype\_embedding} (the characteristics of the prototype) and a \emph{prototype\_value} (the canonical forecast curve of the prototype). An adapter module projects the input window's embedding into a lower-dimensional ``key'', which is compared against every prototype's \emph{prototype\_embedding} via softmax routing, and the
final forecast is the score-weighted sum of all \emph{prototype\_value} curves.

A core design choice of ProtoTS is that the prototypes form a tree.
Training starts with 6--12 root prototypes, and a splitting mechanism is triggered whenever the validation loss stops improving for too long. At that point, the highest-error prototypes --- ranked by cumulative loss normalized by activation count --- each split into several children, so that each child specializes on one aspect of the parent's more complex scenario.

\subsection{Motivations from ProtoTS}

Building on ProtoTS's design, our analysis identifies three aspects
that motivate the design choices in ConceptTS:

\textbf{(1) Randomly initialized prototypes.}
Because prototypes are initialized randomly and shaped by gradient
descent, their semantic meaning is not specified by the design itself.
Interpreting a trained prototype typically requires a domain expert to
inspect its activation patterns and the corresponding
\emph{prototype\_value} shape after training, which can be
time-consuming and sometimes ambiguous.

\textbf{(2) Encoder that mixes time and feature axes.}
ProtoTS's encoder blends information along both the time and feature
axes before the prototype module. This design is effective for
capturing joint temporal-channel patterns, but it makes it harder to
isolate specific cross-channel semantics --- a concept such as
\textit{``extremely low temperature leads to an electricity-load
spike''} would be difficult to recover from a single learned prototype
through post-hoc inspection alone.

\textbf{(3) Auxiliary path through value\_bias (most impactful for our
design).}
In addition to the softmax-normalized activation score --- which
represents \textit{``how similar the input window is to this
prototype''} --- ProtoTS multiplies each prototype's contribution by a
\emph{value\_bias} scalar produced by a separate MLP that reads from
the encoder output embedding. The final forecast is therefore
\[
    \hat{y} = \sum_{\text{prototypes}}
    \text{activation\_score} \times \text{value\_bias} \times
    \text{prototype\_value}.
\]
In our experiments we observed that, since the encoder is co-trained
with the prototype layer, the model tends to converge toward a
configuration in which prototype activation scores become nearly
uniform across input windows, while most of the input-dependent
forecasting signal is carried by the
\emph{value\_bias $\times$ prototype\_value} channel. Because
\emph{value\_bias} reads directly from the encoder output, this
pathway does not preserve the same interpretability structure as the
activation channel. ProtoTS includes an entropy-based regularization
term designed to keep activation scores non-uniform; we found that its
default weight had limited effect on our data, and when we raised the
weight enough for the entropy term to actively shape training,
forecast accuracy dropped noticeably --- suggesting that in this
configuration the trained model's accuracy depends in part on the
auxiliary pathway remaining active.

Fig~\ref{fig:protots_activation} illustrates such problem intuitively.

\subsection{From Prototypes to Concepts}

In the early stage of this project we tried to adapt ProtoTS's structure directly to improve on the three aspects above, but every attempt failed to maintain accuracy while restoring interpretability. In retrospect, this makes sense: constraining each prototype to a single canonical forecast curve is too restrictive for multivariate time-series forecasting, where a ``prototype'' would need to jointly represent a target curve \emph{and} every meaningful cross-channel condition that
could produce it.

We also considered attaching multiple \emph{prototype\_value} curves to each prototype to increase expressive capacity, but that raises a new problem: how many prototypes are needed to cover every realistic scenario? Even a broad prototype such as \textit{``summer night''} has many variants along orthogonal axes --- \textit{``rainy or not''}, \textit{``off-work rush hour or late
night''}, \textit{``uncomfortably hot or tolerably hot''} --- and
enumerating every combination scales exponentially. ProtoTS's
tree-splitting mechanism partially addresses this by growing new
prototypes on demand, but each split adds parameters and complicates training, and the required prototype count still grows
combinatorially with the number of feature axes.

This bottleneck is what led us to the concept-based framework. The
seemingly unbounded number of prototypes required for a real
multivariate task can be replaced by a much smaller set of concepts
whose on/off combinations effectively span the same space. Each
concept focuses on one or two feature channels, which makes both
training and post-hoc inspection substantially easier, while
expressive capacity remains high: even 20--30 well-chosen concepts
can combine to describe tens of thousands of distinct scenarios ---
the same coverage prototype-based methods can only achieve through
uncontrolled tree growth.

\section{Concept Failure Mode Analysis}

\begin{table*}[!ht]
\centering
\caption{Per-concept test-set AUROC across the three ConceptTS
bottlenecks. Concepts are sorted by AUROC within each bottleneck.
Forecast sub-window AUROC is averaged over all sub-window.
\textbf{Bold} entries mark the three concepts with the lowest AUROC in
each bottleneck.}
\label{tab:concept-auroc-full}
\footnotesize
\renewcommand{\arraystretch}{1.05}
\begin{tabular}{@{}lc@{\hskip 40pt}lc@{}}
\toprule
Concept & AUROC & Concept & AUROC \\
\midrule
\multicolumn{4}{l}{\textbf{Lookback concepts (19)}} \\
\midrule
\texttt{is\_history\_winter}             & 1.000 & \texttt{is\_history\_cold}                 & 1.000 \\
\texttt{is\_history\_summer}             & 1.000 & \texttt{is\_history\_humid}                & 0.999 \\
\texttt{is\_history\_high\_so2}          & 0.999 & \texttt{is\_history\_dry}                  & 0.998 \\
\texttt{is\_history\_high\_co}           & 0.998 & \texttt{is\_history\_high\_ozone}          & 0.998 \\
\texttt{is\_history\_rainy}              & 0.997 & \texttt{is\_history\_high\_no2}            & 0.994 \\
\texttt{is\_history\_pm25\_rising}       & 0.992 & \texttt{is\_history\_low\_ozone}           & 0.992 \\
\texttt{is\_history\_low\_wind}          & 0.987 & \texttt{is\_history\_high\_pm25}           & 0.985 \\
\texttt{is\_history\_pm25\_volatile}     & 0.985 & \texttt{is\_history\_high\_pressure}       & 0.966 \\
\textbf{\texttt{is\_history\_wind\_anticorrelated}} & \textbf{0.855} & \textbf{\texttt{is\_history\_pm25\_noisy\_spectrum}} & \textbf{0.797} \\
\textbf{\texttt{is\_history\_pressure\_rising}}      & \textbf{0.676} & & \\
\midrule
\multicolumn{4}{l}{\textbf{Forecast sub-window concepts (47), AUROC averaged over 16 sub-windows}} \\
\midrule
\texttt{is\_cold\_block}                 & 1.000 & \texttt{is\_very\_humid\_block}            & 1.000 \\
\texttt{is\_humid\_block}                & 0.999 & \texttt{is\_dry\_block}                    & 0.992 \\
\texttt{is\_high\_co\_block}             & 0.989 & \texttt{is\_high\_so2\_block}              & 0.988 \\
\texttt{is\_hot\_block}                  & 0.987 & \texttt{is\_high\_no2\_block}              & 0.982 \\
\texttt{is\_low\_pm10\_block}            & 0.981 & \texttt{is\_high\_pm10\_block}             & 0.979 \\
\texttt{is\_low\_so2\_block}             & 0.978 & \texttt{is\_winter\_block}                 & 0.977 \\
\texttt{is\_low\_co\_block}              & 0.975 & \texttt{is\_summer\_block}                 & 0.970 \\
\texttt{is\_low\_ozone\_block}           & 0.970 & \texttt{is\_low\_no2\_block}               & 0.960 \\
\texttt{is\_high\_pressure\_block}       & 0.959 & \texttt{is\_high\_ozone\_block}            & 0.957 \\
\texttt{is\_late\_night\_block}          & 0.957 & \texttt{is\_nighttime\_block}              & 0.955 \\
\texttt{is\_daytime\_block}              & 0.951 & \texttt{is\_midday\_block}                 & 0.951 \\
\texttt{is\_evening\_rush\_block}        & 0.949 & \texttt{is\_very\_high\_pressure\_block}   & 0.941 \\
\texttt{is\_low\_pressure\_block}        & 0.931 & \texttt{is\_windy\_block}                  & 0.925 \\
\texttt{is\_warming\_block}              & 0.924 & \texttt{is\_rainy\_block}                  & 0.903 \\
\texttt{is\_spring\_block}               & 0.893 & \texttt{is\_calm\_block}                   & 0.864 \\
\texttt{is\_ozone\_rising\_block}        & 0.848 & \texttt{is\_ozone\_falling\_block}         & 0.844 \\
\texttt{is\_very\_calm\_block}           & 0.829 & \texttt{is\_moderate\_temp\_block}         & 0.808 \\
\texttt{is\_cooling\_block}              & 0.789 & \texttt{is\_autumn\_block}                 & 0.783 \\
\texttt{is\_co\_rising\_block}           & 0.769 & \texttt{is\_pm10\_rising\_block}           & 0.763 \\
\texttt{is\_humidity\_falling\_block}    & 0.761 & \texttt{is\_no2\_rising\_block}            & 0.756 \\
\texttt{is\_pressure\_rising\_block}     & 0.749 & \texttt{is\_stable\_wind\_block}           & 0.716 \\
\texttt{is\_humidity\_rising\_block}     & 0.713 & \texttt{is\_pressure\_falling\_block}      & 0.704 \\
\textbf{\texttt{is\_wind\_rising\_block}} & \textbf{0.700} & \textbf{\texttt{is\_stable\_temp\_block}} & \textbf{0.688} \\
\textbf{\texttt{is\_wind\_falling\_block}} & \textbf{0.684} & & \\
\midrule
\multicolumn{4}{l}{\textbf{Global-forecast concepts (19)}} \\
\midrule
\texttt{is\_forecast\_cold\_regime}          & 1.000 & \texttt{is\_forecast\_humid\_regime}       & 1.000 \\
\texttt{is\_forecast\_dry\_regime}           & 0.999 & \texttt{is\_forecast\_high\_pm10\_regime}  & 0.999 \\
\texttt{is\_forecast\_high\_co\_regime}      & 0.999 & \texttt{is\_forecast\_high\_ozone\_regime} & 0.999 \\
\texttt{is\_forecast\_high\_no2\_regime}     & 0.997 & \texttt{is\_forecast\_humidity\_rising}    & 0.987 \\
\texttt{is\_forecast\_hot\_regime}           & 0.979 & \texttt{is\_forecast\_high\_pressure\_regime} & 0.967 \\
\texttt{is\_forecast\_wind\_rising}          & 0.896 & \texttt{is\_forecast\_has\_rain}           & 0.893 \\
\texttt{is\_forecast\_wind\_falling}         & 0.850 & \texttt{is\_forecast\_stable\_pressure}    & 0.784 \\
\texttt{is\_forecast\_cooling}               & 0.743 & \texttt{is\_forecast\_temp\_large\_swing}  & 0.713 \\
\textbf{\texttt{is\_forecast\_pressure\_falling}} & \textbf{0.704} & \textbf{\texttt{is\_forecast\_pressure\_rising}} & \textbf{0.667} \\
\textbf{\texttt{is\_forecast\_warming}}      & \textbf{0.599} & & \\
\bottomrule
\end{tabular}
\end{table*}

Table~\ref{tab:concept-auroc-full} reports all concepts proposed by the LLM proposer and their test-set AUROC. AUROC measures the probability that a randomly chosen positive example is assigned a higher predicted probability than a randomly chosen negative example, and ranges from $0.5$ (random guessing) to $1.0$ (perfect separation). In each of the three bottlenecks, the three concepts with the lowest AUROC are marked using bold texts.

A clear pattern emerges: the model consistently struggles with concepts that describe \emph{temporal trends or fluctuation patterns}.
Concepts such as \texttt{is\_forecast\_warming}, \texttt{is\_forecast\_pressure\_rising}, and \texttt{is\_stable\_temp\_block} all fall well below the AUROC of
level-based concepts (e.g., \texttt{is\_history\_dry}, \texttt{is\_cold\_block}, which routinely score above $0.9$).

We attribute this failure mode to a fundamental limitation of our
concept-labeling pipeline. Recall that the ground-truth binary label
for each concept is produced at LLM-proposal time by applying a Python
predicate over per-segment summary statistics (mean, standard
deviation, slope, range, etc.). While these statistics are sufficient
for describing concepts based on absolute levels or spreads
(e.g., ``\emph{cold window}'', ``\emph{high humidity}''), they cannot
cleanly express the temporal shape of a curve --- whether it monotonically rises, monotonically falls, fluctuates around a mean, or
follows a more complex trajectory such as a rise-then-fall pattern.
The ground-truth labels for these shape-dependent concepts are
therefore internally inconsistent, giving the model a noisier and more
challenging learning target than for the statistically well-defined
concepts.

A natural remedy, which we leave to future work, is to enrich the
per-segment statistical summary with shape descriptors so that the LLM-generated Python predicates can express shape-based conditions as precisely as level-based ones.

\section{ModernTCN Encoder Structure}

Each ModernTCN block consists of the following modules:

\begin{itemize}
    \item $\mathbf{DWConvolution}$: the tensor is reshaped to $(B, C\times D, N)$, and a depthwise convolution layer $ \text{Conv1D}(\mathit{CD}, \mathit{CD}, \text{kernel}=\kappa, \text{padding}=\kappa/2, \text{groups}=\mathit{CD})$ is applied with a large kernel size $\kappa$ ($\kappa$ = 51 in our setting), performing a temporal mixing per (variable, feature). This step aims to capture the temporal patterns and also preserves the semantic meaning.

    \item $\mathbf{PWConvolution_{1}}$: a grouped pointwise inverted-bottleneck $ \text{Conv1D}(\mathit{CD}, r \cdot \mathit{CD}, 1, \text{groups}=C) \rightarrow \text{GELU} \rightarrow \text{Conv1D}(r \cdot \mathit{CD}, \mathit{CD}, 1, \text{groups}=C) $, aiming to mix information along feature axis on a per-variable basis

    \item $\mathbf{PWConvolution_{2}}$: a grouped pointwise inverted-bottleneck $ \text{Conv1D}(\mathit{DC}, r \cdot \mathit{DC}, 1, \text{groups}=D) \rightarrow \text{GELU} \rightarrow \text{Conv1D}(r \cdot \mathit{DC}, \mathit{DC}, 1, \text{groups}=D) $, aiming to mix information along variable axis on a per-feature basis
    
\end{itemize}

\section{Loss Function Composition}

We describe each of the seven loss terms in Eq.~(8) below:

\paragraph{Forecast Loss.}
$\mathcal{L}_{\mathrm{forecast}}$ is the mean $L_1$ error between
$\widehat{\mathbf{Y}}_{L+1:L+H}$ and $\mathbf{Y}_{L+1:L+H}$ on the z-scored
target. This is the standard loss term that drives the model's forecast
curve towards the ground-truth target.

\paragraph{Concept Loss.}
$\mathcal{L}_{\mathrm{concept}}^{\mathrm{lb}}$,
$\mathcal{L}_{\mathrm{concept}}^{\mathrm{fc}}$, and
$\mathcal{L}_{\mathrm{concept}}^{\mathrm{gfc}}$ denote the concept losses
for the look-back, forecast sub-window, and global forecast bottlenecks,
respectively. Each is the binary cross-entropy between the predicted
concept logits and the LLM-derived binary labels, averaged over concepts
(and, for the sub-window bottleneck, over the $N_{\mathrm{sw}}$ sub-window
positions).

\paragraph{Boundary Smoothness Loss.}
Since the decoder predicts the target sub-window by sub-window and
concatenates the results, the transitions at sub-window boundaries can be
discontinuous. We penalize the $L_1$ difference between adjacent forecast
timesteps at the $N_{\mathrm{sw}}-1$ boundary positions:
\begin{equation}
    \mathcal{L}_{\mathrm{boundary}}
    = \frac{1}{N_{\mathrm{sw}} - 1} \sum_{k=1}^{N_{\mathrm{sw}} - 1}
    \left|\hat{y}_{L + k T_{\mathrm{sw}}} - \hat{y}_{L + k T_{\mathrm{sw}} - 1}\right|.
\end{equation}

\paragraph{Concept Collapse Loss.}
The CEM gate $\hat{\mathbf{c}}_c = p_c\, \mathbf{e}_c^{+} + (1 - p_c)\, \mathbf{e}_c^{-}$
degenerates if $p_c$ becomes near-constant across the batch: the resulting
$\hat{\mathbf{c}}_c$ is nearly identical for every input, so the concept
carries no per-sample information and the decoder cannot use it. Although
the concept losses already penalize wrong probability assignments, we add
an explicit variance-based penalty to prevent this collapse:
\begin{equation}
    \mathcal{L}_{\mathrm{collapse}}
    = \sum_c \max\!\left(0,\, \sigma_{\min} - \mathrm{Var}_{\mathrm{batch}}(p_c)\right),
\end{equation}
where $\sigma_{\min}$ is the minimum variance threshold. Any concept whose
batch-wise activation variance falls below $\sigma_{\min}$ incurs a
positive penalty proportional to how far below the threshold it is.

\paragraph{Residual Regularization Loss.}
As discussed in the model architecture section, a low-dimensional
residual channel is added alongside the concept bottleneck to compensate
for the loss of expressive capacity introduced by the static concept
embeddings. To prevent the decoder from relying too heavily on this
channel and bypassing the concept bottleneck, we add an $L_2$ penalty on
the residual vector:
\begin{equation}
    \mathcal{L}_{\mathrm{residual}}
    = \left\lVert \mathbf{r} \right\rVert_2^{\,2},
\end{equation}
where $\mathbf{r}$ is the residual channel output. The weight
$\alpha_{\mathrm{res}}$ directly controls the trade-off between forecast
accuracy and interpretability.

\section{Experimental Setup Details}

This appendix specifies the details in all the experiments introduced in Experiment Section.

\subsection{Dataset and Chronological Split}

We use the Beijing Multi-Site Air Quality dataset~\cite{chen2019beijingair}
at three monitoring stations: \textit{Aotizhongxin}, \textit{Dingling},
and \textit{Tiantan}. Each station is treated as an independent dataset
and the model is trained separately per station. The raw data spans 1 March 2013 through 28 February 2017 at hourly resolution.

We adopt a strictly chronological split to prevent leakage from future
to past, with training set, validation set, and test set each containing 70\%, 10\%, and 20\% of the data. Missing values are filled via linear interpolation, applied across the full data before splitting.

Global normalization statistics ($\mu$, $\sigma$) used for z-scoring the
target and covariates are estimated only from the training split. The
per-cluster statistics used to condition the LLM proposer and the concept positive-rate filter are also computed on the training split only.

Sliding windows are extracted with look-back length $L = 96$ hours and
forecast horizon $H = 96$ hours, using a stride of 6 hours.

Although training uses a sliding-window stride of 1 hour, the LLM
concept-labeling predicates are evaluated at a coarser stride of 6
hours to make the entire concept proposing stage more efficient. Concretely, for any group of six consecutive training segments
whose look-back windows begin within the same 6-hour block, we compute
the binary concept labels once and reuse them across all six segments.
This design has two motivations. First, it reduces the number of predicate evaluations by $6\times$, keeping the offline labeling pass tractable when the training set contains hundreds of thousands of windows. Second, and more importantly, it matches the natural temporal coherence of the proposed concepts: most of them (e.g., \texttt{is\_history\_cold}, \texttt{is\_forecast\_humid}, \texttt{is\_winter\_block}) describe conditions that persist for at least several hours, so their true binary states rarely flip within a 6-hour window.

\subsection{Model Architecture Hyperparameters}
\begin{itemize}
    \item \textbf{Encoder} (ModernTCN-style): per-channel embedding
    dimension $D = 32$; patchification patch size $P = 4$ and stride
    $S = 4$; depthwise-convolution kernel size $\kappa = 51$;
    ConvFFN inverted-bottleneck expansion ratio $r = 2$; number of
    ModernTCN blocks $= 4$; post-projection output embedding
    dimension $D_{\mathrm{enc}} = 128$. Two independent encoders with
    identical hyperparameters are used, one over the look-back window
    and one over the forecast horizon.
    \item \textbf{Attention pool}: additive-attention MLP hidden
    dimension $= 32$.
    \item \textbf{CEM bottlenecks}: concept embedding dimension
    $d_c = 16$; each concept's scoring head is a single linear
    projection (no hidden layer) from $D_{\mathrm{enc}}$ to a scalar
    logit, followed by a sigmoid. Concept counts per bottleneck used
    in the main results are: lookback $K_{\mathrm{lb}} = 20$;
    forecast sub-window $K_{\mathrm{sw}} = 50$ (separate predictor
    weights per sub-window, shared concept embeddings across
    sub-windows); global forecast $K_{\mathrm{gfc}} = 20$.
    \item \textbf{BiGRU}: input summary dimension $d_{\mathrm{summ}} = 64$; hidden dimension per direction $= 32$ (concatenated output dimension $= 64$); number of layers $= 1$.
    \item \textbf{Sub-decoder}: $B_{\mathrm{dec}} = 3$ residual MLP
    blocks with feedback of the running forecast into blocks $b > 1$;
    per-block MLP hidden dimension $= 512$; sub-window positional
    embedding dimension $d_{\mathrm{id}} = 16$; residual channel
    dimension $= 4$, projected from the look-back encoder's
    global-pooled feature by a single learned linear layer and
    concatenated into the decoder input.
\end{itemize}

\subsection{Loss Weights and Training Procedure}

The seven weighting coefficients in Eq.~(8) are set as follows:
$\alpha_f = 1.0$, $\alpha_{\mathrm{lb}} = 1.0$,
$\alpha_{\mathrm{fc}} = 1.0$, $\alpha_{\mathrm{gfc}} = 1.0$,
$\alpha_{\mathrm{bd}} = 0.01$, $\alpha_{\mathrm{cc}} = 0.05$ (with
variance threshold $\sigma_{\min} = 0.05$), and
$\alpha_{\mathrm{res}} = 0.01$ (default; swept over $\{0.01, 0.1, 1.0\}$
in Table~III). All runs use the Adam optimizer with initial learning
rate $3 \times 10^{-4}$, weight decay $1 \times 10^{-4}$, batch size
$16$, and no learning-rate scheduler (constant learning rate throughout
training). Models are trained for up to $60$ epochs with early
stopping on validation MAE (patience $= 5$ epochs); the checkpoint
with the lowest validation MAE is reported.

\subsection{Baseline Implementations}

Time-Series-Library baselines (Informer, iTransformer, TimeXer,
Crossformer, DLinear) use the official implementations from Time-Series-Library with their recommended default hyperparameters, adjusted only for our look-back length $L=96$ and forecast horizon $H=96$. Tree-based baselines (Random Forest, XGBoost, LightGBM) use the scikit-learn, xgboost, and lightgbm library. TFT, DeepAR, and NHiT use the PyTorch Forecasting implementation. LSTM is a 2-layer, hidden dimension 128 baseline implemented from scratch. All baselines are given identical inputs to ConceptTS in each setting: full future covariates in the future-aware setting, look-back inputs only in the look-back-only setting, in order to make the comparison fair.

\subsection{Experimental Configuration}

For the main forecasting-performance experiments, we set the number of
sub-windows to $N_{\mathrm{sw}} = 24$, which we found offers the best
trade-off between forecast accuracy and model parameter count. At each
station, every model (ConceptTS and all baselines) is trained three times with different random seeds, so we can report both mean performance and run-to-run stability. Concept proposals are generated once per station, and the same concept sets are reused across all three ConceptTS training runs at that station.

For the residual-weight ablation ($\alpha_{\mathrm{res}}$), we fix both
the concept set and the three random seeds. This ensures that any variation across values of $\alpha_{\mathrm{res}}$ reflects the ablation itself rather than concept-proposal or seed noise.

For the case study subsection, we reduce the number of sub-windows to $N_{\mathrm{sw}} = 16$. Fewer sub-windows means each sub-window covers more timesteps, which makes the per-sub-window forecast curves and concept activations easier to read and analyze.

\section{Additional Ablation Studies}

\begin{figure*}[!t]
    \centering
    \begin{subfigure}[b]{0.48\textwidth}
        \centering
        \includegraphics[width=\textwidth]{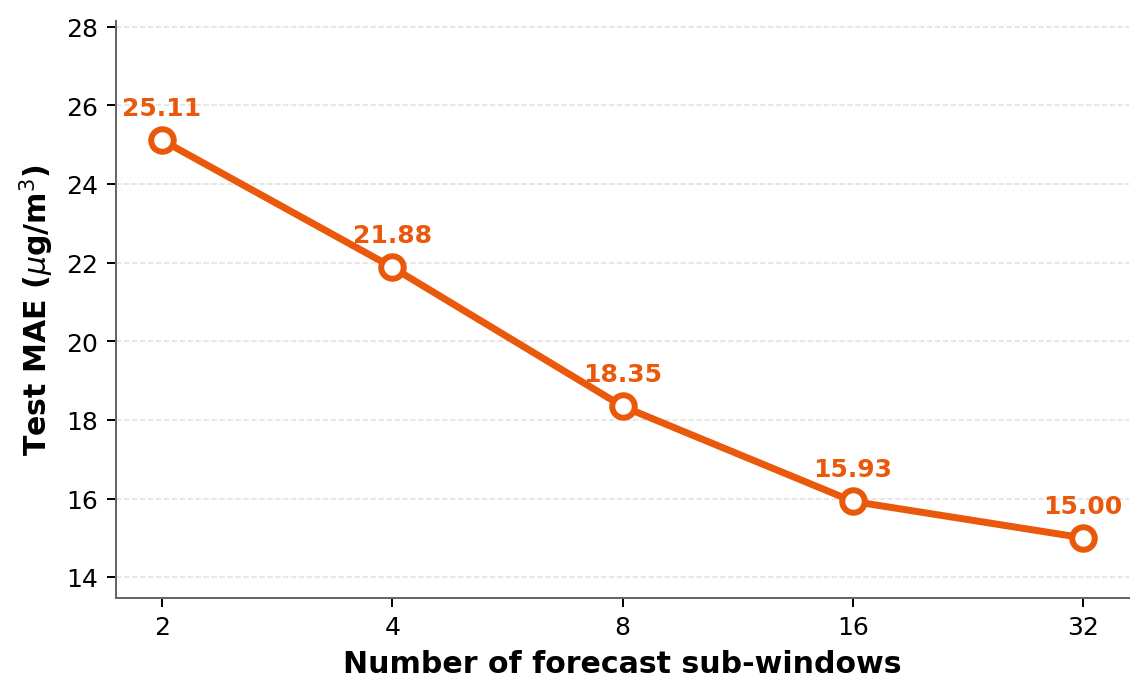}
        \caption{MAE}
        \label{fig:subwindow-mae}
    \end{subfigure}
    \hfill
    \begin{subfigure}[b]{0.48\textwidth}
        \centering
        \includegraphics[width=\textwidth]{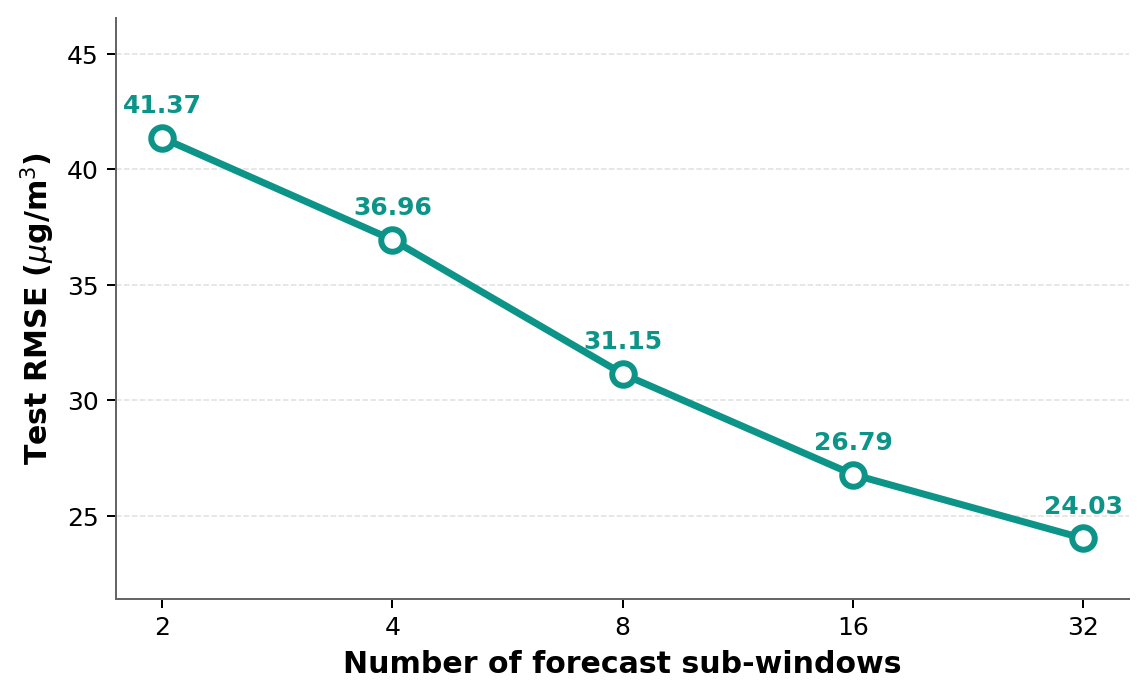}
        \caption{RMSE}
        \label{fig:subwindow-rmse}
    \end{subfigure}
    \caption{Effect of the number of forecast sub-windows on
    forecasting performance. Both metrics decrease monotonically as
    the forecast horizon is partitioned into more sub-windows, with
    diminishing returns beyond $N_{\mathrm{sw}} = 16$.}
    \label{fig:subwindow-sweep}
\end{figure*}

\subsection{Concept Bottleneck Contribution}
To understand how much the forecast depends on each concept bottleneck, we design an ablation that measures the contribution of the three bottlenecks---the lookback bottleneck, the forecast sub-window bottleneck, and the global forecast bottleneck--- directly from a single trained model. We first train the model in the standard setting, then evaluate it on the test set four times: once in the normal configuration, and three additional times, each of which ``blocks out'' one bottleneck. To block out a bottleneck, we replace the activation score of every concept in it with that concept's mean activation over the test set. Since these values are constant across inputs, the corresponding concept context no longer carries any input-dependent information, so the bottleneck is effectively removed from the forecast while the rest of the model, including the other two bottlenecks, is left unchanged. Using the per-concept mean rather
than an arbitrary constant such as zero keeps the decoder input within its normal range, so the measured change in accuracy reflects the loss of concept information rather than the effect of feeding the model an unusual input.

We adopt this test-time procedure, instead of retraining the model with a bottleneck removed each time, for a specific reason. If a bottleneck were removed before training, the model could compensate by re-routing the same information through the remaining concepts, which would mask the true role of the removed bottleneck. By intervening only at test time on a fixed trained model, we avoid such compensation and obtain a direct estimate of how much each bottleneck contributes to the final prediction. We report the test error under each configuration in Table~\ref{tab:bottleneck_ablation}: a large increase in error when a
bottleneck is blocked indicates that the model relies heavily on it, whereas a small change indicates that the model's forecast doesn't get influenced significantly by it.

\begin{table}[t]
    \centering
    \caption{Ablation study of the three concept bottlenecks in ConceptTS on Aotizhongxin. $\Delta$ MAE reports the absolute performance degradation, in terms of MAE, when compared to the full model. All values are in $\mu\mathrm{g}/\mathrm{m}^3$.}
    \label{tab:bottleneck_ablation}
    \renewcommand{\arraystretch}{1.2}
    \begin{tabular}{lcc}
        \toprule
        Configuration & MAE & $\Delta$ MAE \\
        \midrule
        Full model                                & 15.042 & --- \\
        Remove lookback bottleneck                & 15.079 & $\downarrow 0.037$ \\
        Remove global forecast bottleneck         & 16.454 & $\downarrow 1.412$ \\
        Remove forecast sub-window bottleneck     & 58.236 & $\downarrow 43.194$ \\
        \bottomrule
    \end{tabular}
\end{table}

The results reveal a clear ordering of importance among the three bottlenecks. Blocking the lookback bottleneck degrades the forecast only marginally ($\downarrow 0.037 \mu\mathrm{g}/\mathrm{m}^3$ MAE), removing the global forecast bottleneck causes a larger but still moderate drop ($\downarrow 1.412 \mu\mathrm{g}/\mathrm{m}^3$ MAE), and blocking the forecast sub-window bottleneck is by far the most damaging ($\downarrow 43.193 \mu\mathrm{g}/\mathrm{m}^3$ MAE), leaving the model with almost no forecasting ability.

This ordering is consistent with our setting and the nature of the task. Because we adopt a future-aware configuration, the model has access to the true covariates over the forecast window, and therefore predicts mainly from what is known about the future rather than from the past. This effect is reinforced by the fact that PM2.5 lacks the strong, regular periodicity of series such as electricity load, so the recent history is only a weak predictor of future values; the lookback concepts are thus used merely as auxiliary information that slightly refines the forecast, which explains their small contribution. The dominant contribution comes from the per-sub-window concept activations: these
concepts describe the conditions within each individual sub-window, and the model needs such fine-grained, temporally resolved information to generate the forecast at every step of the horizon. The global forecast bottleneck plays an intermediate role --- rather than resolving each sub-window, it summarizes the overall regime of the entire forecast window and provides a coarse context that refines the prediction on top of the sub-window details, which is consistent with its moderate but non-negligible effect.

\subsection{Number of Sub-windows}

We further analyze how the number of forecast sub-windows, $N_{\mathrm{sw}}$, affects the overall performance of ConceptTS. To
isolate the effect of this hyperparameter, we disable the residual
channel so that all forecasting must flow through the concept bottlenecks. To keep the comparison fair, we reuse the same LLM-proposed concept sets (look-back, global forecast, and forecast
sub-window) across every training run, and for each value of $N_{\mathrm{sw}}$ we train the model twice with different random seeds and report the average.

The results are shown in Fig.~\ref{fig:subwindow-sweep}. Both MAE and RMSE decrease monotonically as $N_{\mathrm{sw}}$ increases: MAE drops from $25.11$ to $15.00\ \mu\mathrm{g/m}^3$ as $N_{\mathrm{sw}}$ grows from 2 to 32, and RMSE drops from $41.37$ to $24.03\ \mu\mathrm{g/m}^3$ over the same range. This trend is intuitive --- a larger $N_{\mathrm{sw}}$ shortens each sub-window, which lets the corresponding forecast sub-window concepts describe finer-grained local trends and events. However, the improvement is not free: the number of concept-scoring heads and CEM embeddings grows linearly with $N_{\mathrm{sw}}$, increasing both training time and total parameter count. Based on this trade-off, we set $N_{\mathrm{sw}} = 24$ in our main performance experiments, which sits in the region where the accuracy curve has largely flattened but the parameter count remains manageable.

\section{LLM Concept-Proposal Prompt}
\label{app:prompt}

This appendix reproduces the full prompt template that ConceptTS
sends to the concept-proposing LLM, instantiated for the Beijing
PM2.5 forecasting task used throughout the paper. Task-dependent
portions are marked with double braces --- e.g.,
\verb|{{input_length}}|, \verb|{{n_subwindows}}|,
\verb|{{channel_descriptions}}| --- and are substituted at runtime
with the concrete values used for a given task. The template is sent verbatim to the LLM once per proposal call; the JSON output is then parsed into executable Python predicates that label every training segment.

\begin{lstlisting}[style=promptstyle]
You are a domain expert in urban air-quality forecasting
(PM2.5) for Beijing. The dominant drivers of PM2.5
concentration in this region are local emissions plus
meteorological conditions that either disperse or trap
pollutants -- low wind speed (no horizontal dispersion),
high pressure and inversions (no vertical mixing), high
humidity (hygroscopic particle growth), precipitation
(wet scavenging), and chemical interactions with co-
emitted pollutants (PM10, SO2, NO2, CO, O3). You are
helping design a CONCEPT BOTTLENECK MODEL for
interpretable PM2.5 air-quality forecasting.

The model supports two training modes, both using a
single shared LOOKBACK bottleneck but with different
sub-window concept sets:

  * FUTURE-AWARE mode uses 2 + N bottlenecks:
    - 1 LOOKBACK bottleneck (K_lb concepts summarising
      the whole history window).
    - N FORECAST-SUB-WINDOW bottlenecks: K_fc concepts
      applied at each of the N forecast sub-windows,
      sharing the same static concept embeddings but
      with per-sub-window predictor heads.
    - 1 GLOBAL-FORECAST bottleneck (K_gfc concepts
      spanning the whole forecast horizon as a single
      unit).
  * LOOK-BACK-ONLY mode uses 1 + N_lb_sw bottlenecks:
    - 1 LOOKBACK bottleneck (same as above).
    - N_lb_sw LOOKBACK-SUB-WINDOW bottlenecks: K_lb_sw
      concepts applied at each of the N_lb_sw lookback
      sub-windows.

You will propose ALL FOUR concept sets in a single JSON
response. Each training run uses either the forecast
side (FUTURE-AWARE mode) OR the lookback-sub-window
side (LOOK-BACK-ONLY mode); the whole-history LOOKBACK
set is always used. Emitting all four in one shot means
one LLM run supports both training modes.

## Task Context

- Dataset: {{dataset_name}}
- Target: {{target_variable}} ({{target_description}})
- Time resolution: {{time_resolution}}
- Lookback window: {{input_length}} steps
- Forecast window: {{output_length}} steps split into
  {{n_subwindows}} sub-windows of {{subwindow_length}}
  steps each
- Total training segments: {{total_segments}}

## Available Feature Channels

{{channel_descriptions}}

## Global Statistics (across all training segments)

{{global_statistics}}

## Cluster Summaries ({{n_clusters}} K-means regimes)

Use these to pick thresholds that actually split the
data into the regimes you see below.

{{cluster_summaries}}

## Schema 1 -- segment dict (for LOOKBACK concepts)

def matches(seg) -> bool receives a dict with:
  - lookback_stats: per-channel {mean, median, std,
    min, max, range, skewness, kurtosis, slope,
    delta_half, volume} plus FFT-derived
    {fft_dom_period_h, fft_dom_amp, fft_spec_entropy,
    fft_lowfreq_ratio, fft_amp_p{24,12,8,6}h}.
  - cross_channel_stats: target_corr_{PM10, WSPM,
    TEMP, DEWP, PRES, RAIN}.
  - time_context: lookback_start / lookback_end /
    season.

## Schema 2 -- sub-window dict (for FORECAST and
##            LOOKBACK-SUB-WINDOW concepts)

def matches(sub_seg) -> bool receives a dict with:
  - subwindow_idx: int (0..N-1).
  - stats: per-channel shape identical to
    lookback_stats but WITHOUT FFT fields. For
    FORECAST-SUB-WINDOW and FORECAST-GLOBAL concepts,
    PM2.5 is MASKED and MUST NOT be referenced;
    LOOKBACK-SUB-WINDOW concepts MAY use PM2.5
    because the lookback history is observed.
  - time_context: subwindow_idx, start, end, season.
  - cross_channel_stats: inherited from the parent
    segment.

Per-sub-window concepts (FORECAST-SUB-WINDOW and
LOOKBACK-SUB-WINDOW) fire {{n_subwindows}} times per
segment, once per sub-window. GLOBAL-FORECAST
concepts fire ONCE per segment; their sub_seg.stats
and sub_seg.time_context cover the WHOLE
{{output_length}}-step forecast window rather than a
single sub-window.

## Your Task

*** CRITICAL CONSTRAINT ***
PM2.5 IS THE PREDICTION TARGET. The forecast window's
PM2.5 values are masked / unavailable at inference time.
  - LOOKBACK and LOOKBACK-SUB-WINDOW concepts MAY
    reference PM2.5 (the lookback history is observed).
  - FORECAST-SUB-WINDOW concepts MUST NOT reference
    PM2.5.
  - FORECAST-GLOBAL concepts MUST NOT reference PM2.5.

Forecast-side concepts must be built EXCLUSIVELY from
meteorology (TEMP, PRES, DEWP, RAIN, WSPM), other
pollutants (PM10, SO2, NO2, CO, O3), and calendar
(year, month, day, hour, season).

Propose FOUR concept sets:
  * Exactly {{n_lookback_concepts}} LOOKBACK concepts.
  * Exactly {{n_lookback_sub_window_concepts}}
    LOOKBACK-SUB-WINDOW concepts (each rule fires
    {{n_lookback_subwindows}} times per segment, once
    per lookback sub-window).
  * Exactly {{n_forecast_concepts}} FORECAST-SUB-WINDOW
    concepts (each rule fires {{n_subwindows}} times
    per segment, once per forecast sub-window).
  * Exactly {{n_forecast_global_concepts}}
    FORECAST-GLOBAL concepts (each rule fires ONCE per
    segment on the whole {{output_length}}-step
    horizon).

Hard requirements for every function:
  1. Named exactly `matches`. Single positional arg.
  2. ATOMIC: body is one return statement over a single
     boolean expression; helper lookups allowed. No
     loops, no if/else.
  3. Defensive .get(...) access; missing keys must
     return False.
  4. No imports. Standard Python only.
  5. Target positive rate in [5%, 60%] across the
     {{total_segments}} segments (over segment*sub-
     window pairs for sub-window rules).
  6. Diverse categories: season, time_of_day, calendar,
     weather_temperature, weather_other,
     weather_dynamics, pollutant_chemistry,
     cross_channel, regime, frequency (lookback only),
     aggregate / threshold / persistence / trend (good
     for the FORECAST-GLOBAL set), misc.
  7. No compounds (the CBM combines concepts at the
     embedding level).
  8. NO TARGET LEAKAGE: forecast-side rules must not
     touch PM2.5 in any way (name, description, or
     code); such rules will be REJECTED.
  9. MANDATORY SELF-CHECK: scan every forecast-side
     entry before emitting JSON; delete and replace
     any rule that references PM2.5.

## Sub-window trend rules

Short sub-windows ({{subwindow_hours}} hours) have
systematically ~0 OLS slope for cyclic channels (TEMP,
DEWP, wind, pollutants). For "is X rising / falling
in this block" concepts (both FORECAST-SUB-WINDOW and
LOOKBACK-SUB-WINDOW), use stats[X]['delta_half']
(mean(second half) - mean(first half)) instead of
stats[X]['slope']. Use slope only for the LOOKBACK
bottleneck (whole 96-hour window) and FORECAST-GLOBAL
concepts (whole horizon).

## Output Format

Return a single JSON object of the shape:

{
  "lookback_concepts":
      [... {{n_lookback_concepts}} entries ...],
  "lookback_sub_window_concepts":
      [... entries ...],
  "forecast_concepts":
      [... entries ...],
  "forecast_global_concepts":
      [... entries ...]
}

Each entry has fields {id, name, description, category,
function}, where "function" is the executable Python
source for `def matches(...)`. IDs run 0..K-1
independently per array. Return ONLY the JSON object,
no prose before or after.
\end{lstlisting}

\section{Test on Electricity Load Dataset}

\begin{table*}[t]
    \centering
    \caption{Forecasting performance on LOF-PC dataset.
    Values are computed on the $z$-scored target (dimensionless)}
    \label{tab:lof-performance}
    \renewcommand{\arraystretch}{1.15}
    \begin{tabular}{lcccccccc}
        \toprule
        Metric & \sys & Informer & TFT & Random Forest & XGBoost & LightGBM & DeepAR & NHiT \\
        \midrule
        MAE  ($z$) & 0.3502 & 0.3319 & 0.2835 & 0.2387 & 0.1646 & 0.1531 & 0.2598 & 0.1797 \\
        RMSE ($z$) & 0.4917 & 0.4764 & 0.3684 & 0.3324 & 0.2528 & 0.2323 & 0.4059 & 0.2545 \\
        \bottomrule
    \end{tabular}
\end{table*}

We also evaluate our model on the LOF dataset, an electric load
forecasting dataset released by the ProtoTS authors. LOF contains
electricity consumption records from four regions of a northern Chinese province at 15-minute resolution, together with 22 exogenous features covering weather measurements, temporal indicators, and calendar information. We train ConceptTS on the region \textit{PC} and compare it against the same set of baseline models used in the experiment section. Following the original ProtoTS paper, we use the future-aware setting so that all
models have access to the future exogenous features at inference time.

Table~\ref{tab:lof-performance} reports the results. Unlike the Beijing PM2.5 task, ConceptTS trails every baseline model on LOF. We attribute this drop to two factors. First, the LOF dataset shows a noticeable distribution shift between the training and testing sets, and our model handles this shift less gracefully than the tree-based and Transformer baselines. Second, and more importantly, electricity load has strong, clear temporal patterns on both daily and yearly cycles, and our concept bottleneck design is not well suited to capture such long-term periodicities. Every concept in our vocabulary describes short-term local behavior (e.g., \textit{``the sub-window is in the evening peak''}, \textit{``the last hour was hot''}), and none of them can directly express a seasonal or annual pattern. This also explains why ConceptTS remains competitive on the PM2.5 task: air quality
forecasting depends mainly on short-term cross-channel context --- wind speed, other pollutant concentrations, dew point --- which is exactly what our concept vocabulary is designed to expose. Such scope limitation --- concepts describing only short-term local behavior --- is a genuine constraint of our current design. Consequently, equipping ConceptTS with concepts at longer temporal scales (daily, seasonal, annual) is a natural direction for future work.

%% file: references.bib
@article{theissler2022explainable,
  author  = {Theissler, Andreas and Spinnato, Francesco and Schlegel, Udo and Guidotti, Riccardo},
  title   = {Explainable {AI} for Time Series Classification: A Review, Taxonomy and Research Directions},
  journal = {IEEE Access},
  volume  = {10},
  pages   = {100700--100724},
  year    = {2022},
  doi     = {10.1109/ACCESS.2022.3207765},
  url     = {https://doi.org/10.1109/ACCESS.2022.3207765}
}

@misc{zhao2023interpretation,
  author        = {Zhao, Ziqi and Shi, Yucheng and Wu, Shushan and Yang, Fan and Song, Wenzhan and Liu, Ninghao},
  title         = {Interpretation of Time-Series Deep Models: A Survey},
  howpublished  = {arXiv preprint arXiv:2305.14582},
  year          = {2023},
  eprint        = {2305.14582},
  archivePrefix = {arXiv},
  primaryClass  = {cs.LG},
  doi           = {10.48550/arXiv.2305.14582},
  url           = {https://arxiv.org/abs/2305.14582}
}

@article{park2026transparent,
  author    = {Park, Youngjin and Tong, Anh and Lee, Sehyun and Seong, Jihyeon and Xie, Qin and Choi, Jaesik},
  title     = {Towards Transparent Time Series Analysis: Exploring Methods and Enhancing Interpretability},
  journal   = {ACM Computing Surveys},
  volume    = {58},
  number    = {9},
  pages     = {240:1--240:36},
  month     = mar,
  year      = {2026},
  publisher = {Association for Computing Machinery},
  doi       = {10.1145/3794839},
  url       = {https://doi.org/10.1145/3794839}
}

@article{wang2026deep,
  author  = {Wang, Yuxuan and Wu, Haixu and Dong, Jiaxiang and Liu, Yong and Wang, Chen and Long, Mingsheng and Wang, Jianmin},
  title   = {Deep Time Series Models: A Comprehensive Survey and Benchmark},
  journal = {IEEE Transactions on Pattern Analysis and Machine Intelligence},
  pages   = {1--20},
  month   = may,
  year    = {2026},
  doi     = {10.1109/TPAMI.2026.3690845},
  url     = {https://doi.org/10.1109/TPAMI.2026.3690845}
}

@inproceedings{liu2024timexplusplus,
  author    = {Liu, Zichuan and Wang, Tianchun and Shi, Jimeng and Zheng, Xu and Chen, Zhuomin and Song, Lei and Dong, Wenqian and Obeysekera, Jayantha and Shirani, Farhad and Luo, Dongsheng},
  title     = {{TimeX++}: Learning Time-Series Explanations with Information Bottleneck},
  booktitle = {Proceedings of the 41st International Conference on Machine Learning},
  series    = {Proceedings of Machine Learning Research},
  volume    = {235},
  pages     = {32062--32082},
  publisher = {PMLR},
  month     = jul,
  year      = {2024},
  url       = {https://proceedings.mlr.press/v235/liu24bl.html}
}

@article{huang2026sigtime,
  author  = {Huang, Yu-Chia and Chen, Juntong and Liu, Dongyu and Ma, Kwan-Liu},
  title   = {{SigTime}: Learning and Visually Explaining Time Series Signatures},
  journal = {IEEE Transactions on Visualization and Computer Graphics},
  volume  = {32},
  number  = {2},
  pages   = {2099--2113},
  month   = feb,
  year    = {2026},
  doi     = {10.1109/TVCG.2025.3644956},
  url     = {https://doi.org/10.1109/TVCG.2025.3644956}
}

@article{liu2022mtv,
  author    = {Liu, Dongyu and Alnegheimish, Sarah and Zytek, Alexandra and Veeramachaneni, Kalyan},
  title     = {{MTV}: Visual Analytics for Detecting, Investigating, and Annotating Anomalies in Multivariate Time Series},
  journal   = {Proceedings of the ACM on Human-Computer Interaction},
  volume    = {6},
  number    = {CSCW1},
  pages     = {103:1--103:30},
  month     = apr,
  year      = {2022},
  publisher = {Association for Computing Machinery},
  doi       = {10.1145/3512950},
  url       = {https://doi.org/10.1145/3512950}
}

@article{haoyietal-informer-2021,
  author  = {Zhou, Haoyi and Zhang, Shanghang and Peng, Jieqi and Zhang, Shuai and Li, Jianxin and Xiong, Hui and Zhang, Wancai},
  title   = {{Informer}: Beyond Efficient Transformer for Long Sequence Time-Series Forecasting},
  journal = {Proceedings of the AAAI Conference on Artificial Intelligence},
  volume  = {35},
  number  = {12},
  pages   = {11106--11115},
  month   = may,
  year    = {2021},
  doi     = {10.1609/aaai.v35i12.17325}
}

@article{Zeng2022AreTE,
  author  = {Zeng, Ailing and Chen, Muxi and Zhang, Lei and Xu, Qiang},
  title   = {Are {Transformers} Effective for Time Series Forecasting?},
  journal = {Proceedings of the AAAI Conference on Artificial Intelligence},
  volume  = {37},
  number  = {9},
  pages   = {11121--11128},
  month   = jun,
  year    = {2023},
  doi     = {10.1609/aaai.v37i9.26317}
}

@inproceedings{liu2023itransformer,
  author    = {Liu, Yong and Hu, Tengge and Zhang, Haoran and Wu, Haixu and Wang, Shiyu and Ma, Lintao and Long, Mingsheng},
  title     = {{iTransformer}: Inverted Transformers Are Effective for Time Series Forecasting},
  booktitle = {International Conference on Learning Representations},
  pages     = {11116--11140},
  year      = {2024},
  url       = {https://proceedings.iclr.cc/paper_files/paper/2024/hash/2ea18fdc667e0ef2ad82b2b4d65147ad-Abstract-Conference.html}
}

@inproceedings{peng2025protots,
  author    = {Peng, Ziheng and Ren, Shijie and Gu, Xinyue and Yang, Linxiao and Wang, Xiting and Sun, Liang},
  title     = {{ProtoTS}: Learning Hierarchical Prototypes for Explainable Time Series Forecasting},
  booktitle = {International Conference on Learning Representations},
  year      = {2026},
  url       = {https://openreview.net/forum?id=IbcdVwzLrp}
}

@inproceedings{koh2020conceptbottleneckmodels,
  author    = {Koh, Pang Wei and Nguyen, Thao and Tang, Yew Siang and Mussmann, Stephen and Pierson, Emma and Kim, Been and Liang, Percy},
  title     = {Concept Bottleneck Models},
  booktitle = {Proceedings of the 37th International Conference on Machine Learning},
  series    = {Proceedings of Machine Learning Research},
  volume    = {119},
  pages     = {5338--5348},
  publisher = {PMLR},
  year      = {2020},
  url       = {https://proceedings.mlr.press/v119/koh20a.html}
}

@inproceedings{yang2023languagebottlelanguagemodel,
  author    = {Yang, Yue and Panagopoulou, Artemis and Zhou, Shenghao and Jin, Daniel and Callison-Burch, Chris and Yatskar, Mark},
  title     = {Language in a Bottle: Language Model Guided Concept Bottlenecks for Interpretable Image Classification},
  booktitle = {Proceedings of the IEEE/CVF Conference on Computer Vision and Pattern Recognition},
  pages     = {19187--19197},
  month     = jun,
  year      = {2023},
  publisher = {IEEE},
  doi       = {10.1109/CVPR52729.2023.01839}
}

@inproceedings{sun2025conceptbottlenecklargelanguage,
  author    = {Sun, Chung-En and Oikarinen, Tuomas and Ustun, Berk and Weng, Tsui-Wei},
  title     = {Concept Bottleneck Large Language Models},
  booktitle = {International Conference on Learning Representations},
  pages     = {89371--89411},
  year      = {2025},
  url       = {https://proceedings.iclr.cc/paper_files/paper/2025/hash/de4ce91dfe56b919ee1c228d6a78f866-Abstract-Conference.html}
}

@article{lim2020temporalfusiontransformersinterpretable,
  author    = {Lim, Bryan and Ar{\i}k, Sercan {\"O}. and Loeff, Nicolas and Pfister, Tomas},
  title     = {Temporal Fusion Transformers for Interpretable Multi-Horizon Time Series Forecasting},
  journal   = {International Journal of Forecasting},
  volume    = {37},
  number    = {4},
  pages     = {1748--1764},
  month     = oct,
  year      = {2021},
  publisher = {Elsevier},
  doi       = {10.1016/j.ijforecast.2021.03.012}
}

@inproceedings{Panetal2021,
  author    = {Pan, Qingyi and Hu, Wenbo and Chen, Ning},
  title     = {Two Birds with One Stone: Series Saliency for Accurate and Interpretable Multivariate Time Series Forecasting},
  booktitle = {Proceedings of the Thirtieth International Joint Conference on Artificial Intelligence},
  editor    = {Zhou, Zhi-Hua},
  pages     = {2884--2891},
  publisher = {International Joint Conferences on Artificial Intelligence Organization},
  month     = aug,
  year      = {2021},
  doi       = {10.24963/ijcai.2021/397}
}

@inproceedings{espinosa2022cem,
  author    = {Espinosa Zarlenga, Mateo and Barbiero, Pietro and Ciravegna, Gabriele and Marra, Giuseppe and Giannini, Francesco and Diligenti, Michelangelo and Shams, Zohreh and Precioso, Frederic and Melacci, Stefano and Weller, Adrian and Li{\'o}, Pietro and Jamnik, Mateja},
  title     = {Concept Embedding Models: Beyond the Accuracy-Explainability Trade-Off},
  booktitle = {Advances in Neural Information Processing Systems},
  volume    = {35},
  pages     = {21400--21413},
  publisher = {Curran Associates, Inc.},
  year      = {2022},
  doi       = {10.52202/068431-1555}
}

@inproceedings{luo2024moderntcn,
  author    = {Luo, Donghao and Wang, Xue},
  title     = {{ModernTCN}: A Modern Pure Convolution Structure for General Time Series Analysis},
  booktitle = {International Conference on Learning Representations},
  pages     = {31728--31770},
  year      = {2024},
  url       = {https://proceedings.iclr.cc/paper_files/paper/2024/hash/86b1437c1e4c3b3c4debff98234a67e7-Abstract-Conference.html}
}

@article{challu2023nhits,
  author  = {Challu, Cristian and Olivares, Kin G. and Oreshkin, Boris N. and Garza Ramirez, Federico and Mergenthaler Canseco, Max and Dubrawski, Artur},
  title   = {{NHITS}: Neural Hierarchical Interpolation for Time Series Forecasting},
  journal = {Proceedings of the AAAI Conference on Artificial Intelligence},
  volume  = {37},
  number  = {6},
  pages   = {6989--6997},
  month   = jun,
  year    = {2023},
  doi     = {10.1609/aaai.v37i6.25854}
}

@misc{chen2019beijingair,
  author       = {Chen, Song},
  title        = {{Beijing Multi-Site Air Quality}},
  howpublished = {UCI Machine Learning Repository},
  year         = {2017},
  doi          = {10.24432/C5RK5G},
  url          = {https://archive.ics.uci.edu/dataset/501/beijingmultisiteairqualitydata}
}

@article{breiman2001rf,
  author  = {Breiman, Leo},
  title   = {Random Forests},
  journal = {Machine Learning},
  volume  = {45},
  number  = {1},
  pages   = {5--32},
  month   = oct,
  year    = {2001},
  doi     = {10.1023/A:1010933404324}
}

@inproceedings{chen2016xgboost,
  author    = {Chen, Tianqi and Guestrin, Carlos},
  title     = {{XGBoost}: A Scalable Tree Boosting System},
  booktitle = {Proceedings of the 22nd ACM SIGKDD International Conference on Knowledge Discovery and Data Mining},
  pages     = {785--794},
  publisher = {Association for Computing Machinery},
  address   = {New York, NY, USA},
  year      = {2016},
  doi       = {10.1145/2939672.2939785}
}

@inproceedings{ke2017lightgbm,
  author    = {Ke, Guolin and Meng, Qi and Finley, Thomas and Wang, Taifeng and Chen, Wei and Ma, Weidong and Ye, Qiwei and Liu, Tie-Yan},
  title     = {{LightGBM}: A Highly Efficient Gradient Boosting Decision Tree},
  booktitle = {Advances in Neural Information Processing Systems},
  volume    = {30},
  pages     = {3146--3154},
  publisher = {Curran Associates, Inc.},
  year      = {2017}
}

@article{salinas2020deepar,
  author    = {Salinas, David and Flunkert, Valentin and Gasthaus, Jan and Januschowski, Tim},
  title     = {{DeepAR}: Probabilistic Forecasting with Autoregressive Recurrent Networks},
  journal   = {International Journal of Forecasting},
  volume    = {36},
  number    = {3},
  pages     = {1181--1191},
  year      = {2020},
  publisher = {Elsevier},
  doi       = {10.1016/j.ijforecast.2019.07.001}
}

@inproceedings{wang2024timexer,
  author    = {Wang, Yuxuan and Wu, Haixu and Dong, Jiaxiang and Qin, Guo and Zhang, Haoran and Liu, Yong and Qiu, Yunzhong and Wang, Jianmin and Long, Mingsheng},
  title     = {{TimeXer}: Empowering Transformers for Time Series Forecasting with Exogenous Variables},
  booktitle = {Advances in Neural Information Processing Systems},
  volume    = {37},
  pages     = {469--498},
  publisher = {Curran Associates, Inc.},
  year      = {2024},
  doi       = {10.52202/079017-0015}
}

@inproceedings{zhang2023crossformer,
  author    = {Zhang, Yunhao and Yan, Junchi},
  title     = {{Crossformer}: Transformer Utilizing Cross-Dimension Dependency for Multivariate Time Series Forecasting},
  booktitle = {International Conference on Learning Representations},
  year      = {2023},
  url       = {https://openreview.net/forum?id=vSVLM2j9eie}
}

@article{hochreiter1997lstm,
  author  = {Hochreiter, Sepp and Schmidhuber, J{\"u}rgen},
  title   = {Long Short-Term Memory},
  journal = {Neural Computation},
  volume  = {9},
  number  = {8},
  pages   = {1735--1780},
  month   = nov,
  year    = {1997},
  doi     = {10.1162/neco.1997.9.8.1735}
}
